\documentclass[10pt,twocolumn,letterpaper]{article}

\usepackage{cvpr}
\usepackage{times}
\usepackage{epsfig}
\usepackage{graphicx}
\usepackage{amsmath}
\usepackage{amssymb}
\usepackage{booktabs}
\usepackage{colortbl}
\usepackage{xcolor}
\usepackage{subcaption}
\usepackage{float}
\usepackage{multirow}
\usepackage[labelsep=period]{caption}
\RequirePackage{enumitem}
\setlist[itemize]{nosep}

\usepackage[pagebackref=true,breaklinks=true,letterpaper=true,colorlinks,bookmarks=false]{hyperref}

\cvprfinalcopy 

\def\cvprPaperID{844} 

\ifcvprfinal\fi
\begin{document}

\title{TOM-Net: Learning Transparent Object Matting from a Single Image}

\author{Guanying Chen\thanks{indicates equal contribution}, \enspace Kai Han\footnotemark[1], \enspace Kwan-Yee K. Wong\\
The University of Hong Kong, Hong Kong\\
{\tt\small \{gychen, khan, kykwong\}@cs.hku.hk}
}

\maketitle

\begin{abstract}
This paper addresses the problem of transparent object matting. Existing image matting approaches for transparent objects often require tedious capturing procedures and long processing time, which limit their practical use. In this paper, we first formulate transparent object matting as a refractive flow estimation problem. We then propose a deep learning framework, called {\em TOM-Net}, for learning the refractive flow. Our framework comprises two parts, namely a multi-scale encoder-decoder network for producing a coarse prediction, and a residual network for refinement. At test time, TOM-Net takes a single image as input, and outputs a matte (consisting of an object mask, an attenuation mask and a refractive flow field) in a fast feed-forward pass. As no off-the-shelf dataset is available for transparent object matting, we create a large-scale synthetic dataset consisting of 178K images of transparent objects rendered in front of images sampled from the Microsoft COCO dataset. We also collect a real dataset consisting of 876 samples using 14 transparent objects and 60 background images. Promising experimental results have been achieved on both synthetic and real data, which clearly demonstrate the effectiveness of our approach. 
\end{abstract}

\section{Introduction}
\begin{figure}[h!] \begin{center}
    \includegraphics[width=0.5\textwidth]{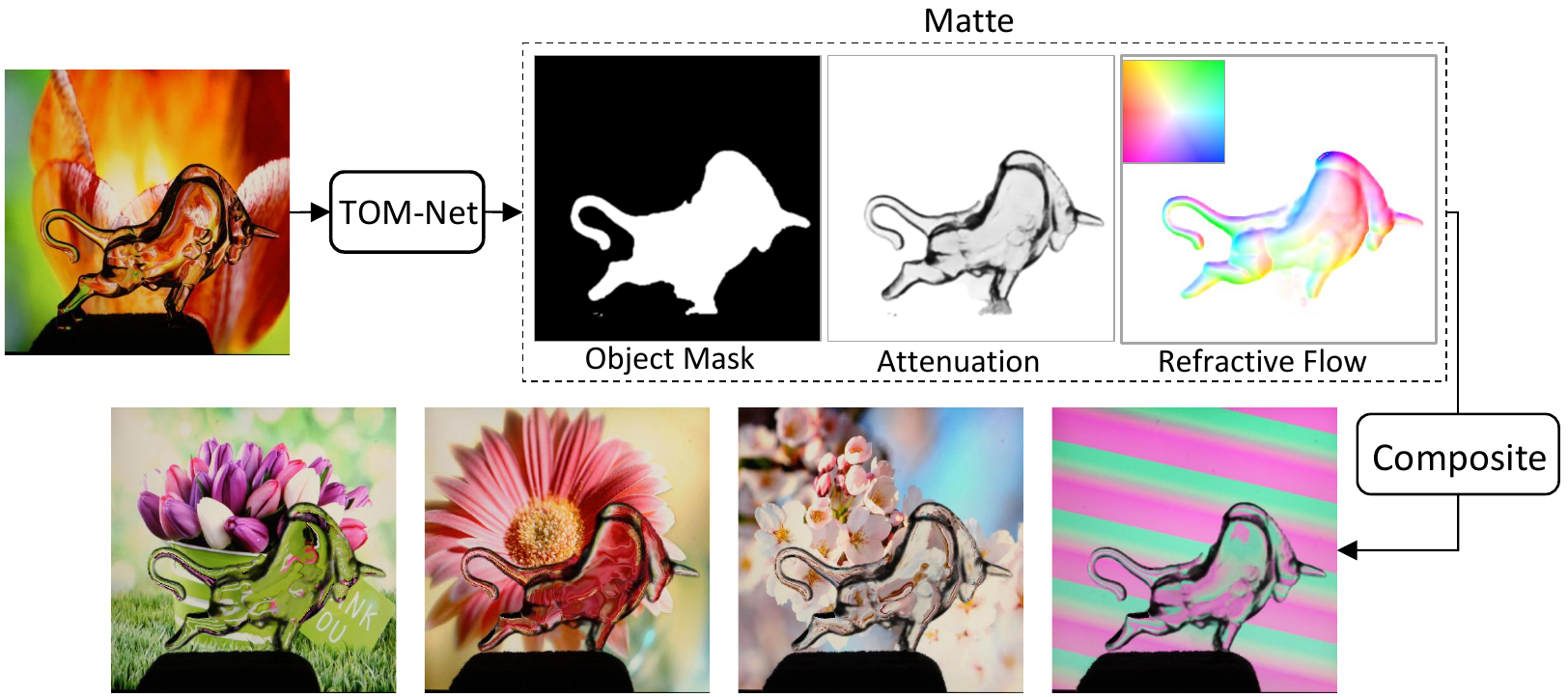}
\end{center}
\caption{Given an image of a transparent object as input, our model can estimate the environment matte (consisting of an object mask, an attenuation mask and a refractive flow field) in a feed-forward pass. The transparent object can then be composited onto new background images with the extracted matte.}
\label{fig:intro}
\end{figure}

Image matting refers to the process of extracting the foreground matte of an image by locating the region of the foreground object and estimating the opacity of each pixel inside the foreground region. The foreground object can then be composited onto a new background image using the \emph{matting equation} \cite{smith1996blue} 
\begin{equation}
    \label{eq:alphamatte}
    C = F + (1-\alpha)B,  \quad \alpha \in [0, 1],
\end{equation}
where $C$ denotes the composited color, $F$ the foreground color, $B$ the background color, and $\alpha$ the opacity.

Image matting has been widely used in image editing and film production. However, most of the existing methods are tailored for opaque objects, and cannot handle transparent objects whose appearances depend on how light is refracted from the background.

To model the effect of refraction, Zongker \etal \cite{zongker1999environment} introduced \emph{environment matting} as 
\begin{equation}
    \label{eq:em_general}
    C = F + (1-\alpha)B + \Phi, \quad \alpha \in [0, 1],
\end{equation}
where $\Phi$ is the contribution of environment light caused by refraction or reflection at the foreground object. Besides estimating the foreground shape, environment matting also describes how objects interact with the background. 

Many efforts \cite{chuang2000environment, wexler2002image, peers2003wavelet, zhu2004frequency, duan2011flexible, duan2015compressive} have been devoted to improving the seminal work of  \cite{zongker1999environment}. The resulting methods often require either a huge number of input images to achieve a higher accuracy, or specially designed patterns to reduce the number of required images. They are in general all very computational expensive.

In this paper, we focus on environment matting for transparent objects. It is highly ill-posed, if not impossible, to estimate an accurate environment matte for transparent objects from a single image with an arbitrary background. Given the huge solution space, there may exist multiple objects and backgrounds that can produce the same refractive effect. In order to make the problem more tractable, we simplify our problem to estimating an environment matte that can produce visually realistic refractive effect from a single image, instead of estimating a highly accurate refractive flow. We define the environment matte in our model as a triple consisting of an object mask, an attenuation mask and a refractive flow field. Realistic refractive effect can then be obtained by compositing the transparent object onto new background images (see Fig.~\ref{fig:intro}). 

Inspired by the great successes of convolutional neural networks (CNNs) in high-level computer vision tasks, we propose a convolutional neural network, called TOM-Net, for simultaneous learning of an object mask, an attenuation mask and a refractive flow field from a single image with an arbitrary background. The key contributions of this paper can be summarized as follows:

\begin{itemize}
  \item We introduce a simple and efficient model for transparent object matting as simultaneous estimation of an object mask, an attenuation mask and a refractive flow field.
  \item We propose a convolutional neural network, TOM-Net, to learn an environment matte of a transparent object from a single image. To the best of our knowledge, TOM-Net is the first CNN that is capable of learning transparent object matting.
  \item We create a large-scale synthetic dataset and a real dataset as a benchmark for learning transparent object matting. Our TOM-Net has produced promising results on both the synthetic and real datasets.
\end{itemize}  

\section{Related Work}
In this section, we briefly review representative works on environment matting and recent works on CNN based image mating. 

\vspace{-1.5em} 
\paragraph{Environment matting}
\label{par:Environment Matting}
Zongker \etal\cite{zongker1999environment} introduced the concept of environment matting, and assumed each foreground pixel being originated from a single rectangular region of the background. They obtained the environment matte by identifying the corresponding background region for each foreground pixel using three monitors and multiple images. Chuang \etal\cite{chuang2000environment} extended \cite{zongker1999environment} in two ways. First, they replaced the single rectangular supporting area for a foreground pixel with multiple 2D oriented Gaussian strips. This makes it possible for their method to model the effects of color dispersion, multiple mapping and glossy reflection. Second, they simplified the environment matting equation by assuming the object colorless and specular transparent. This allows them to achieve real time environment matting (RTEM). The environment matte was then extracted with one image taken in front of a pre-designed pattern. However, RTEM requires background images to segment the transparent objects, and depends on a time-consuming off-line processing. Wexler \etal \cite{wexler2002image} introduced a probabilistic model based method which assumes each background point has a probability to make contribution towards the color of a certain foreground point. Their approach does not require pre-designed patterns during data acquisition, but it still needs multiple images and can only model thin transparent objects. Peers and Dutr{\'e} \cite{peers2003wavelet} used a large number of wavelet basis backgrounds to obtain the environment matte, and their method can also model the effect of diffuse reflection. Based on the fact that a signal can be decomposed uniquely in the frequency domain, Zhu and Yang \cite{zhu2004frequency} proposed a frequency-based approach to extract an accurate environment matte. They used Fourier analysis to solve the decomposition problem. Both \cite{peers2003wavelet} and \cite{zhu2004frequency} require a large number of images to extract the matte (e.g., \cite{peers2003wavelet} needs 2,400 images and \cite{zhu2004frequency} needs 4,096 images for an image of size $1024\times 1024$), making them not very practical. Recently, compressive sensing theory has been applied to environment matting to reduce the number of images required. Duan \etal \cite{duan2011fast} applied this theory in  the spatial domain and Qian \etal \cite{qian2015frequency} applied it in the frequency domain. However, the number of images needed is still in the order of hundreds. In contrast, our work can estimate an environment matte from a single image in a fast feed-forward computation without the need for pre-designed patterns or additional background images.

Yeung \etal \cite{yeung2011matting} proposed an interactive way to estimate an environment matte given an image containing a transparent object. Their method requires users to manually mark the foreground and background in the image, and models the refractive effect using a thin-plate-spline transformation. Their method does not produce an accurate environment matte, but instead a visually pleasing refractive effect. Our method shares the same spirit, but does not involve any human interaction.

\vspace{-1.5em} 
\paragraph{CNN based image matting}
Although the potential of CNN on transparent object matting has not yet been explored, some existing works have adopted CNNs for solving the general image matting problem. Shen \etal \cite{shen2016deep} introduced a CNN for image matting of color portrait images. Cho \etal \cite{cho2016natural} proposed a network to predict a better alpha matte by taking the matting results of the traditional method and normalized color image as input. Xu \etal \cite{xu2017deep} introduced a deep learning framework that can estimate an alpha matte from an image and its trimap. However, none of these methods can be applied directly to the task of transparent object matting as object opacity alone is not sufficient to model the refractive effect. 

\begin{figure*}[htb] \centering
    \includegraphics[width=\textwidth]{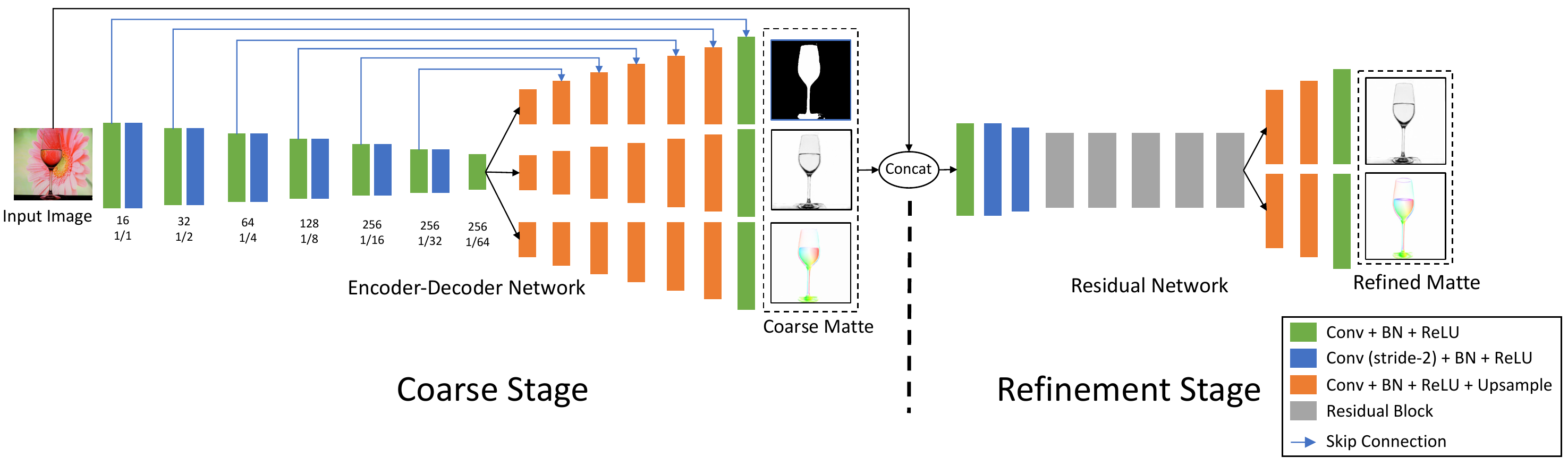}
    \caption{TOM-Net architecture. The left subnetwork is the CoarseNet and the right subnetwork is the RefineNet. (Best viewed in color.)} \label{fig:networkStructure}
\end{figure*}

\section{Matting Formulation}
\label{sec:Transparent Object Matting with Refractive Flow}
As a transparent object may have multiple optical properties (e.g., color attenuation, translucency and reflection), estimating an accurate environment matte for a generic transparent object from a single image is very challenging.  
Following the work of \cite{chuang2000environment}, we cast environment matting to a refractive flow estimation problem by assuming that each foreground pixel only originates from one point in the background due to refraction. Compared to the seminal work of \cite{zongker1999environment}, which models each foreground pixel as a linear combination of a patch in the background, our formulation is more tractable and can be easily encoded using a CNN.

In \cite{zongker1999environment}, the per-pixel environment matting is obtained through leveraging color information from multiple background images. Given a set of pre-designed background patterns, matting is formulated as
\begin{equation}
    \label{eq:em_origin}
    C = F + (1-\alpha)B + \sum_{i=1}^{m} R_i \mathcal{M}(\mathbf{T}_i, \mathbf{A}_i),
\end{equation}
where $F$, $B$ and $\alpha$ denote the foreground color, background color and weight, respectively. The last term in (\ref{eq:em_origin}) accounts for the environment light accumulated from $m$ pre-designed background images ($m=3$ in \cite{zongker1999environment}). $R_i$ is a factor describing the contribution of light emanating from the $i$-$th$ background image $\mathbf{T}_i$. $\mathcal{M}(\mathbf{T}_i, \mathbf{A}_i)$ denotes the average color of a rectangular region $\mathbf{A}_i$ on the background image $\mathbf{T}_i$. 

To obtain an environment matte, the transparent object is placed in front of the monitor(s), and multiple pictures of the object are captured with the monitor(s) displaying different background patterns\footnote{For an image of size $512\times 512$, 18 pictures and around 20 minutes processing time are needed.}. Generally, a surface point receives light from multiple directions, especially for a diffuse surface. When it comes to a (perfectly) transparent object, however, a surface point will only receive light from one direction as determined by the law of refraction. Consider a single background image as the only light source, the problem can be modeled as
\begin{equation}
    \label{eq:em_simplify1}
    C = F + (1-\alpha)B + R \mathcal{M}(\mathbf{T}, P),
\end{equation}
where $\mathcal{M}(\mathbf{T}, P)$ is a bilinear sampling operation at location $P$ on the background image $\mathbf{T}$. Further, by assuming a colorless transparent object, we have $F = 0$ and $R$ becomes a  light attenuation index $\rho$ (a scalar value). The formulation in (\ref{eq:em_simplify1}) can be simplified to
\begin{equation}
    \label{eq:em_simplify2}
    C = (1-\alpha)B + \rho \mathcal{M}(\mathbf{T}, P),
\end{equation}
where $\rho \in [0, 1]$ denotes the attenuation index.

Here, we use refractive flow to model the refractive effect of a transparent object. The refractive flow of a foreground pixel is defined as the offset between the foreground pixel and its refraction correspondence on the background image. 

We further introduce a binary foreground mask to define the object region in the image. The matting equation can now be rewritten as
\begin{equation}
    \label{eq:em_simplify3}
    C = (1 - m) B + m\rho \mathcal{M}(\mathbf{T}, P),
\end{equation}
where $m \in\{0, 1\}$ denotes background ($m = 0$) or foreground ($m = 1$). The matte can then be estimated by solving $m$, $\rho$ and $P$ for each pixel in the input image containing the transparent object\footnote{For each observed pixel, we have 7 unknowns ($3$ for $B$, $2$ for $P$, $1$ for $m$ and $1$ for $\rho$).}.

\section{Learning Transparent Object Matting}
\label{sec:Method}
In this section, we present a two-stage deep learning framework, called TOM-Net, for learning transparent object matting (see Fig.~\ref{fig:networkStructure}). The first stage, denoted as CoarseNet, is a multi-scale encoder-decoder network that takes a single image as input, and predicts an object mask, an attenuation mask and a refractive flow field simultaneously. CoarseNet is capable of predicting a robust object mask. However, the estimated attenuation mask and refractive flow field lack local structural details. 
To overcome this problem, we introduce the second stage of TOM-Net, denoted as RefineNet, to achieve a sharper attenuation mask and a more detailed refractive flow field. RefineNet is a residual network \cite{he2016deep} that takes both the input image and the output of CoarseNet as input. After training, our TOM-Net can predict an environment matte from a single image in a fast feed-forward pass.

\subsection{Encoder-Decoder for Coarse Prediction}
\label{sub:Encoder-decoder Net for Coarse Prediction}
The first stage of our TOM-Net (i.e., CoarseNet) is based on mirror-link CNN introduced in \cite{shi2016learning}. Mirror-link CNN was proposed to learn non-lambertian object intrinsic decomposition. Its output consists of an albedo map, a shading map and a specular map. It shares a similar output structure with our transparent object matting task (i.e., three output branches sharing the same spatial dimensionality). Therefore, it is reasonable for us to adapt mirror-link CNN for our CoarseNet. 

The mirror-link CNN adapted for our CoarseNet consists of one shared encoder and three distinct decoders. The encoder contains six down-sampling convolutional blocks, leading to a down-sampling factor of $64$ in the bottleneck layer. Features in the encoder layers are connected to the decoder layers having the same spatial dimensions through skip connections \cite{ronneberger2015u}. Cross-links \cite{shi2016learning} are introduced to make different decoders share the same input in each layer, so that decoders can better utilize the correlation between different predictions.

Learning with multi-scale loss has been proved to be helpful in dense prediction tasks (e.g., \cite{eigen2014depth,fischer2015flownet}). Since we formulate the problem of transparent object matting as refractive flow estimation, which is a dense prediction task, we augment our mirror-link CNN with multi-scale loss similar to \cite{fischer2015flownet}. 
We use four different scales in our model, where the first scale starts from the decoder features with a down-sampling factor of 8 and the largest scale has the same spatial dimensions as the input.

In contrast to the recent two stage framework for image matting \cite{xu2017deep}, our TOM-Net has a shared encoder and three parallel decoders to accommodate different outputs. Also, we augment our CoarseNet with multi-scale loss and cross-link. Moreover, TOM-Net is trained from scratch while the encoder in \cite{xu2017deep} is initialized with the pre-trained VGG16.


\subsection{Loss Function for Coarse Stage}
\label{sub:Loss Function for CoarseNet}
CoarseNet takes a single image as input and predicts the environment matte as a triple consisting of an object mask, an attenuation mask and a refractive flow field. The learning of CoarseNet is supervised by the ground-truth matte using an object mask segmentation loss $\mathcal{L}_{ms}$, attenuation regression loss $\mathcal{L}_{ar}$, and refractive flow regression loss $\mathcal{L}_{fr}$. Besides, the predicted matte is expected to render an image as close to the input image as possible when applied to the ground-truth background. Hence, in addition to the supervision of the matte, we also take image reconstruction loss $\mathcal{L}_{ir}$ into account. Note that the ground-truth background is only used to calculate the reconstruction error during training but not needed during testing. CoarseNet can therefore be trained by minimizing 
\begin{align}
    \mathcal{L}^c = \alpha^c_{ms} \mathcal{L}_{ms} + \alpha^c_{ar} \mathcal{L}_{ar} + \alpha^c_{fr} \mathcal{L}_{fr} + \alpha^c_{ir} \mathcal{L}_{ir},
\end{align}
where 
$\alpha^c_{ms}, \alpha^c_{ar}, \alpha^c_{fr}, \alpha^c_{ir}$ are weights for the corresponding loss terms. 

\vspace{-1.3em} 
\paragraph{Object mask segmentation loss}
\label{par:Object Mask Classification Loss}
Object mask segmentation is simply a spatial binary classification problem. The output of the object mask decoder has a dimension of $2\times H\times W$, where $H$ and $W$ denote the height and width of the input. We normalize the output with {\em softmax} and compute the loss using the binary cross-entropy function
\begin{equation}
    \mathcal{L}_{ms} = -\frac{1}{HW} \sum_{ij} (\tilde{M}_{ij}\log(P_{ij}) + (1-\tilde{M}_{ij}) \log(1-P_{ij})),
\end{equation}
where $\tilde{M}_{ij} \in \{0,1\}$ and $P_{ij}\in [0,1]$ represent ground truth and normalized foreground probability of the pixel at $(i, j)$, respectively.

\vspace{-1.3em} 
\paragraph{Attenuation regression loss} 
The predicted attenuation mask has a dimension of  $1\times H\times W$. The value of this mask is in the range of $[0, 1]$, where $0$ indicates no light can pass and $1$ indicates the light will not be attenuated. 
We adopt a mean square error (MSE) loss
\begin{equation}
    \mathcal{L}_{ar} = \frac{1}{HW} \sum_{ij} (A_{ij}-\tilde{A}_{ij})^2,
\end{equation}
where $A_{ij}$ is the predicted attenuation index and $\tilde{A}_{ij}$ the ground truth at $(i, j)$.
\vspace{-1.3em} 
\paragraph{Refractive flow regression loss}
\label{par:Disparity Smoothness Loss}
The predicted refractive flow field has a dimension of $2\times H\times W$, where we have one channel for the horizontal displacement and another for the vertical displacement. We normalize the refractive flow with $tanh$ activation and multiply it by the width of the input, such that the output is constrained in the range of $[-W, W]$.
We adopt an average end-point error (EPE) loss
\begin{equation}
    \mathcal{L}_{fr} = \frac{1}{HW} \sum_{ij} \sqrt{(F^x_{ij}-\tilde{F}^x_{ij})^2 + (F^y_{ij}-\tilde{F}^y_{ij})^2},
\end{equation}
where $(F^x, F^y)$ and $(\tilde{F}^x, \tilde{F}^y)$ denote the predicted flow and the ground truth, respectively.

\vspace{-1.3em} 
\paragraph{Image reconstruction loss}
\label{par:Image Reconstruction Loss}
We use MSE loss to measure the dissimilarity between the reconstructed image and the input image. 
Denoting the reconstructed image by $I$ and the ground-truth image (i.e., the input image) by $\tilde{I}$, the reconstruction loss is given by
\begin{equation}
    \mathcal{L}_{ir} = \frac{1}{HW} \sum_{ij} \Vert I_{ij}-\tilde{I}_{ij}\Vert_2^2.
\end{equation}

\vspace{-1.3em} 
\paragraph{Implementation details}
\label{par:Implementation Details}
In all experiments, we empirically set $\alpha^c_{ms}=0.1, \alpha^c_{ar}=1, \alpha^c_{fr}=0.01,$ and $\alpha^c_{ir}=1$. The loss weights for different scales are $\frac{1}{2^{(4 -\text{s})}}$, where $s \in\{1,2,3,4\}$ denotes the scale. 
CoarseNet contains $8M$ parameters and it takes about 2.5 days to train with Adam optimizer \cite{kingma2014adam} on a single GPU.  We first train the CoarseNet from scratch until convergence and then train the RefineNet. 

\subsection{Residual Learning for Matte Refinement}
\label{par:Residual Learning for Matte Refinement}
As the attenuation mask and the refractive flow field predicted by the CoarseNet lack structural details, a refinement stage is needed to produce a detailed matte. Observing that residual learning is particularly suitable for tasks whose input and output are largely similar \cite{kim2016accurate,Nah_2017_CVPR}, we propose a residual network, denoted as RefineNet, to refine the matte predicted by the CoarseNet. 

We concatenate the input image and the output of the CoarseNet to form the input of the RefineNet. As the object mask predicted by the CoarseNet is already plausible, the RefineNet only outputs an attenuation mask and a refractive flow field. The parameters of the CoarseNet are fixed when training the refinement stage. 

\vspace{-1.3em} 
\paragraph{Loss for the refinement stage}
\label{par:Loss for Refinement}
The overall loss for the refinement stage is
\begin{align}
    \mathcal{L}^r = \alpha^r_{ar} \mathcal{L}_{ar} + \alpha^r_{fr} \mathcal{L}_{fr} ,
\end{align}
where $\mathcal{L}_{ar}$ is the refinement attenuation regression loss, $\mathcal{L}_{fr}$ the refinement flow regression loss,  and $\alpha^r_{ar}$, $\alpha^r_{fr}$ their weights. The definitions of these two losses are identical to those defined in the first stage. 
We found that adding the image reconstruction loss in the refinement stage did reduce the image reconstruction error during training, but was not helpful in reserving sharp edges of the refractive flow field (e.g., mouth of a glass), which is essential to maintain the details of an object. Therefore, we  remove the image reconstruction loss here. The
$tanh$ activation for refractive flow is also omitted in this stage to encourage the network to focus on boundary regions that may have larger prediction errors.

\vspace{-1.3em} 
\paragraph{Implementation details}
\label{par:Implementation Details}
We set $\alpha^r_{ar}=1$, $\alpha^r_{fr}=1$ for the refinement. RefineNet contains $1M$ parameters and it takes about 2 days to train with Adam optimizer. RefineNet is randomly initialized during training.

\section{Dataset for Learning and Evaluation}
\label{sub:Dataset}
As no off-the-shelf dataset for transparent object matting is available, and it is very tedious and difficult to produce a large real dataset with ground-truth object masks, attenuation masks and refractive flow fields, we created a large-scale synthetic dataset by using \emph{POV-Ray} \cite{povray} to render images of synthetic transparent objects. Besides, we also collected a real dataset for evaluation. We will show that our TOM-Net trained on the synthetic dataset can generalize well to real world objects, demonstrating its good transferability.

\subsection{Synthetic Dataset} \label{sub:Synthetic Dataset}
We used a large number of background images and 3D models to render our training samples. We randomly changed the pose of the models, as well as the viewpoint and focal length of the camera in the rendering process to avoid overfitting to a fixed setting.

\vspace{-1.3em} 
\paragraph{Backgrounds Images}
\label{par:Backgrounds Images}
We employed two types of background images, namely scene images and synthetic patterns. For scene images, we randomly sampled images from the Microsoft COCO \cite{lin2014microsoft} dataset\footnote{Other large-scale datasets like ImageNet \cite{deng2009imagenet} can also be used.}. The background images for the training set are sampled from COCO Train2014 and Test2015, while that for the validation set are from COCO Val2014, giving rise to 100K scene images in total. 
For synthetic patterns, we rendered 40K patterns of size $512\times 512$ using \emph{POV-Ray} built-in textures. 

\vspace{-1.3em} 
\paragraph{Transparent Objects}
\label{par:Transparent Object}
We divided common transparent objects into four categories, namely glass, glass with water, lens, and complex shape (see Fig.~\ref{fig:qual_synth} for examples). We constructed parametric 3D models for the first three categories, and generated a large number of models using random parameters. For complex shapes, we constructed parametric 3D models for basic shapes like sweeping-spheres and squashed surface of revolution (SOR) parts, and composed a larger number of models using these basic shapes. We generated 178K 3D models in total, with each model assigned a random refractive index $\lambda \in [1.3, 1.5]$. The distribution of these models in four categories is shown in Tab.~\ref{tab:synth} (first two rows). 

\begin{table}[h] \centering
    \caption{Statistics of the introduced datasets.}
    \label{tab:synth}
    \resizebox{0.40\textwidth}{!}{
    \Large
    \begin{tabular}{c|*{6}{c}}
        \toprule
        Type & Glass & Glass \& Water & Lens & Complex & Total \\
        \midrule
        Synthetic Train & 52K & 26K & 20K & 80K & 178K\\
        Synthetic Val   & 250 & 250 & 200 & 200 & 900\\
        \midrule
        Real Test      & 470 & 103 & 61  & 242 & 876 \\
        \bottomrule
    \end{tabular}
    }
\end{table}

\vspace{-1.3em} 
\paragraph{Ground-truth Matte Generation}
\label{par:Ground Truth Generation}
We obtained the ground-truth object mask of a model by rendering it in front of a black background image and setting its color to white. Similarly, we obtained the ground-truth attenuation mask of a model by simply rendering it in front of a white background image. Finally, we obtained the ground-truth refractive flow field (see Fig.~\ref{fig:qual_synth}) of a model by rendering it in front of a sequence of Gray-coded patterns.

\vspace{-1.3em} 
\paragraph{Data Augmentation}
\label{par:}
To improve the diversity of the training data and narrow the gap between real and synthetic data, extensive data augmentation was carried out on-the-fly. 
For an image with a size of $512\times 512$, we randomly performed color (brightness, contrast and saturation) augmentation (in a range of [-0.2, 0.2]), image scaling (in a range of [0.875, 1.05]), noise perturbation (in a range of [-0.05, 0.05]), and horizontal/vertical flipping. Besides, we also blurred the object boundary to make the synthetic data visually more natural. A patch with a size of $448\times 448$ was then randomly cropped from an augmented image and used as input to train CoarseNet. To speed up the training and save memory, a smaller patch with a size of $384\times 384$ was used to train TOM-Net after the training of CoarseNet.

\vspace{-0.2em} 
\subsection{Real Dataset}
\label{sub:Real Dataset}
To validate the transferability of TOM-Net, we introduce a real dataset, which was collected using 14 objects\footnote{The objects consist of 7 glasses, 1 lens and 6 complex objects. Glasses with water are implicitly included.} and 60 background images, resulting in a dataset of 876 images. The data distribution is summarized in Tab.~\ref{tab:synth} (last row). During the data capturing process, the objects were placed under different poses, with the distances between the camera, object and background uncontrolled. Fig.~\ref{fig:real_qualitative} (second column) shows some sample images from the real dataset.  Note that we do not have the ground-truth matte for the real dataset. We instead captured images of the backgrounds without the transparent objects to facilitate evaluation. 

\vspace{-0.5em} 
\section{Experiments and Results}
\label{sec:Experiemnts and Results}
In this section, we present experimental results and analysis.
Currently, it is non-trivial to have a fair comparison with the previous methods, since none of the them can compute the matte from a single image of a transparent object, and there exists no common datasets and measurements for evaluation.
We performed ablation study for TOM-Net, and evaluated our approach on both synthetic and real data. In addition, a user study was conducted to validate the realism of TOM-Net composites. 
Our code, model and datasets will be made available online: \url{https://guanyingc.github.io/TOM-Net}.

\begin{table}
    \centering
    \caption{Ablation study results. F, A, I, and M are short for flow, attenuation, image reconstruction, and object mask, respectively. (The first value for EPE is measured on the whole image and the second measured within the object region. A-MSE and I-MSE are computed on the whole image.)
    }
    \begin{minipage}{0.41\textwidth}
    \resizebox{\textwidth}{!}{
        \begin{tabular}{*{6}{>{\Large}c}}
        \toprule
        ID & Model Variants & \cellcolor{red!25}F-EPE  & \cellcolor{red!25}A-MSE & \cellcolor{red!25}I-MSE &  \cellcolor{blue!25}M-IoU \\ 
        \midrule
        0 & Background                         & 6.5 / 41.0 & 1.58 & 0.87 & 0.15 \\
        1 & CoarseNet - ($\mathcal{L}^c_{fr}$) & 3.9 / 26.5 & 0.24 & 0.23 & 0.98 \\
        2 & CoarseNet - ($\mathcal{L}^c_{ir}$) & 2.3 / 15.7 & 0.25 & 0.22 & 0.98 \\
        3 & CoarseNet - (multi-scale)          & 2.4 / 16.6 & 0.69 & 0.25 & 0.94 \\
        4 & CoarseNet - (cross-link)           & 2.5 / 17.2 & 0.30 & 0.21 & 0.97  \\
        5 & CoarseNet                          & \textbf{2.2 / 15.4} & 0.28 & 0.18 & 0.97 \\
        \midrule
        6 & CoarseNet + RefineNet              & 2.0 / 13.7 & 0.25 & 0.19 & 0.97 \\
        \bottomrule
    \end{tabular}
    }
    \end{minipage}
    \hspace{-0.7em}
    \begin{minipage}{0.06\textwidth}
    \resizebox{\textwidth}{!}{
    \begin{tabular}{cccc}
        \midrule
        MSE ($\cdot10^{-2}$) \\ \cellcolor{red!25} $\downarrow$ better \\ \cellcolor{blue!25} $\uparrow$ better \\
        \midrule
    \end{tabular}
    }
    \end{minipage}
    \label{tab:self_compare}
\end{table}

\begin{figure} \centering
    \makebox[0.09\textwidth]{\footnotesize Input} 
    \makebox[0.09\textwidth]{\footnotesize Coarse Flow} 
    \makebox[0.09\textwidth]{\footnotesize Refined Flow} 
    \makebox[0.09\textwidth]{\footnotesize Coarse Att.} 
    \makebox[0.09\textwidth]{\footnotesize Refined Att.} 
    \\
    \includegraphics[width=0.090\textwidth]{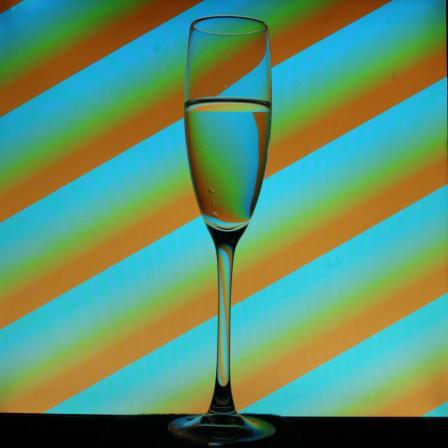}
    \includegraphics[width=0.090\textwidth]{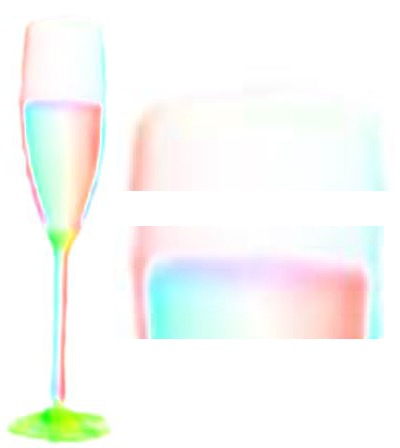}
    \includegraphics[width=0.090\textwidth]{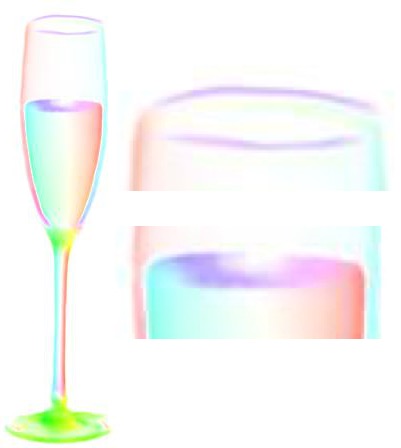}
    \includegraphics[width=0.090\textwidth]{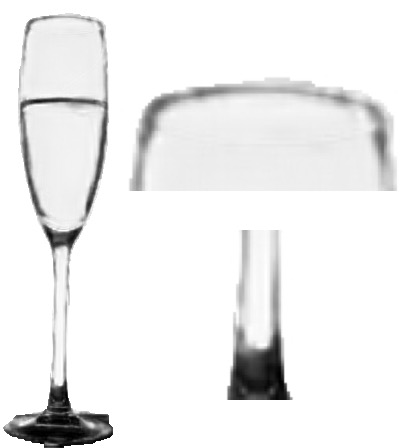}
    \includegraphics[width=0.090\textwidth]{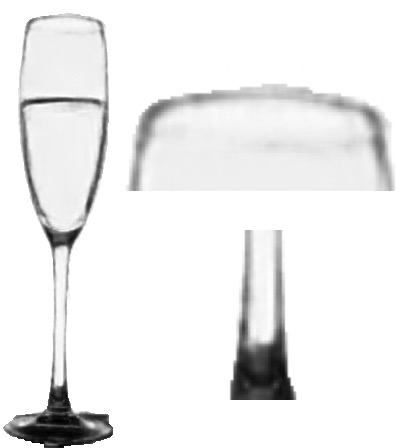}
    \\
    \includegraphics[width=0.090\textwidth]{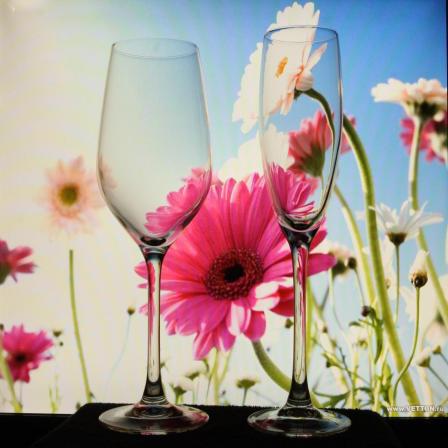}
    \includegraphics[width=0.090\textwidth]{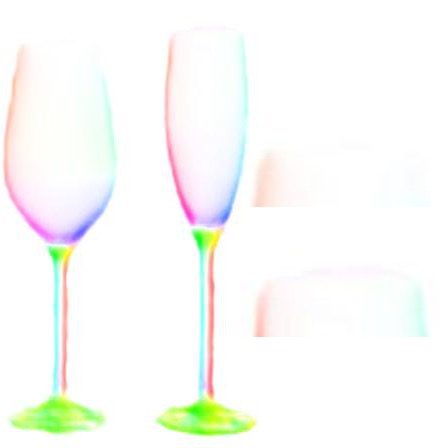}
    \includegraphics[width=0.090\textwidth]{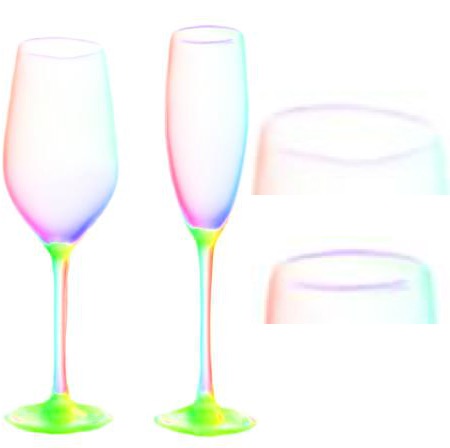}
    \includegraphics[width=0.090\textwidth]{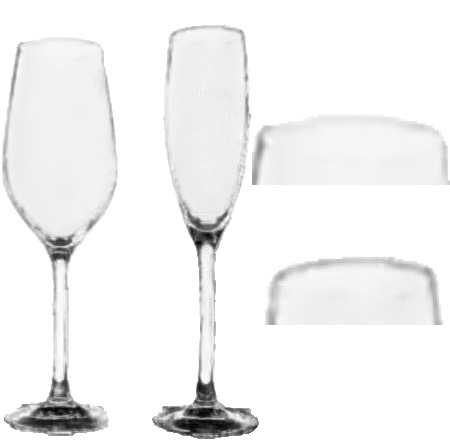}
    \includegraphics[width=0.090\textwidth]{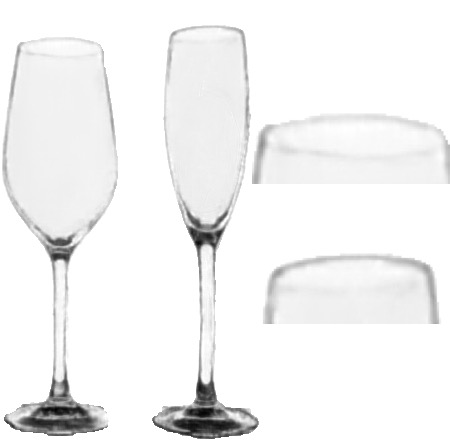}
    \\
    \caption{Visualization of the effectiveness of the refinement stage on real data. After refinement, the refractive flow and attenuation mask have more clear structural details (e.g., glass mouth).} \label{fig:refine}
\end{figure}

\begin{table*} \centering
    \caption{Quantitative results on the synthetic validation set. (The first value for EPE is measured on the whole image and the second measured within the object region. A-MSE and I-MSE are computed on the whole image.)}
    \begin{minipage}{0.93\textwidth}
    \resizebox{\textwidth}{!}{
    \Huge
        \begin{tabular}{c|*{4}{c}|*{4}{c}|*{4}{c}|*{4}{c}|*{4}{c}}
        \toprule
        \multirow{2}{*}{} & \multicolumn{4}{c}{Glass} 
                               & \multicolumn{4}{c}{Glass with Water} 
                               & \multicolumn{4}{c}{Lens} 
                               & \multicolumn{4}{c}{Complex Shape}  
                               & \multicolumn{4}{c}{Average}  \\
                               & \cellcolor{red!25} F-EPE & \cellcolor{red!25}A-MSE & \cellcolor{red!25}I-MSE & \cellcolor{blue!25} M-IoU 
                               & \cellcolor{red!25} F-EPE & \cellcolor{red!25}A-MSE & \cellcolor{red!25}I-MSE & \cellcolor{blue!25} M-IoU 
                               & \cellcolor{red!25} F-EPE & \cellcolor{red!25}A-MSE & \cellcolor{red!25}I-MSE & \cellcolor{blue!25} M-IoU 
                               & \cellcolor{red!25} F-EPE & \cellcolor{red!25}A-MSE & \cellcolor{red!25}I-MSE & \cellcolor{blue!25} M-IoU 
                               & \cellcolor{red!25} F-EPE & \cellcolor{red!25}A-MSE & \cellcolor{red!25}I-MSE & \cellcolor{blue!25} M-IoU \\
        \midrule
        Background    & 3.6 / 30.3 & 1.33 & 0.48 & 0.12 & 6.4 / 53.2 & 1.54 & 0.68 & 0.12 & 10.3 / 39.2 & 1.94 & 1.57 & 0.24 & 6.8 / 56.8 & 2.50 & 0.85 & 0.11 & 6.8 / 44.9 & 1.83 & 0.90 & 0.15 \\
        CoarseNet     & 2.1 / 15.8 & 0.22 & 0.14 & 0.97 & 3.1 / 23.5 & 0.31 & 0.23 & 0.97 & 2.0 / 6.7   & 0.17 & 0.28 & 0.99 & 4.5 / 34.4 & 0.38 & 0.33 & 0.92 & 2.9 / 20.1 & 0.27 & 0.24 & 0.96 \\
        \midrule
        TOM-Net       & 1.9 / 14.7 & 0.21 & 0.14 & 0.97 & 2.9 / 21.8 & 0.30 & 0.22 & 0.97 & 1.9 / 6.6   & 0.15 & 0.29 & 0.99 & 4.1 / 31.5 & 0.37 & 0.32 & 0.92 & 2.7 / 18.6 & 0.26 & 0.24 & 0.96 \\
        \bottomrule
    \end{tabular}
    }
    \end{minipage}
    \hspace{-0.6em}
    \begin{minipage}{0.06\textwidth}
        \resizebox{\textwidth}{!}{
        \begin{tabular}{cc}
            \midrule
            MSE ($\cdot10^{-2}$) \\
            \cellcolor{red!25} $\downarrow$ better \\ 
            \cellcolor{blue!25} $\uparrow$ better\\
            \midrule
        \end{tabular}
        }
    \end{minipage}
    \label{tab:quant_synth}
\end{table*}

\begin{figure*} \centering
    \makebox[0.095\textwidth]{\footnotesize Background} 
    \makebox[0.095\textwidth]{\footnotesize Input} 
    \makebox[0.095\textwidth]{\footnotesize Rec. Image} 
    \makebox[0.095\textwidth]{\footnotesize Rec. Error} 
    \makebox[0.19\textwidth]{\footnotesize Refractive Flow (GT / Est.)} 
    \makebox[0.19\textwidth]{\footnotesize Object Mask (GT / Est.)} 
    \makebox[0.19\textwidth]{\footnotesize Attenuation Mask (GT / Est.)} 
    \\
    \includegraphics[width=0.095\textwidth]{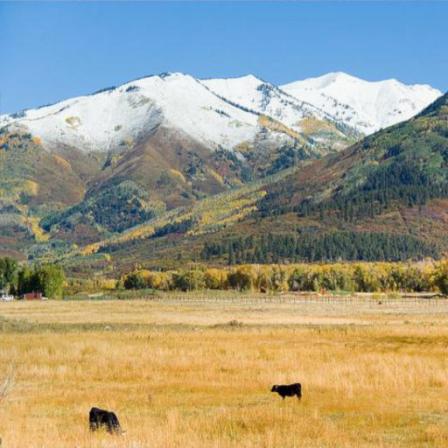}
    \includegraphics[width=0.095\textwidth]{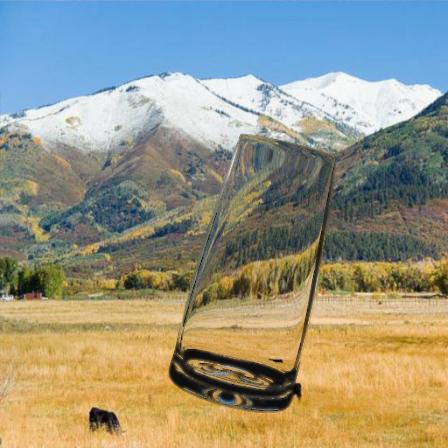}
    \includegraphics[width=0.095\textwidth]{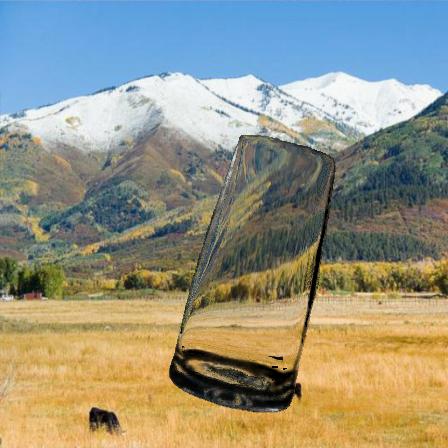}
    \includegraphics[width=0.095\textwidth]{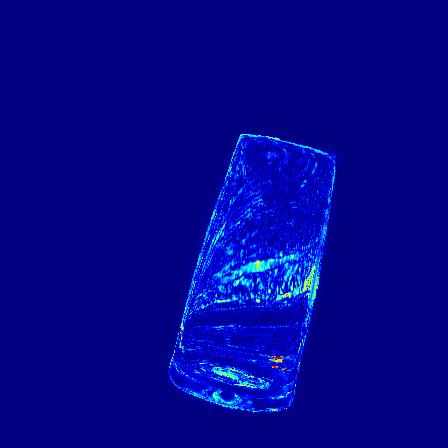}
    \includegraphics[width=0.095\textwidth]{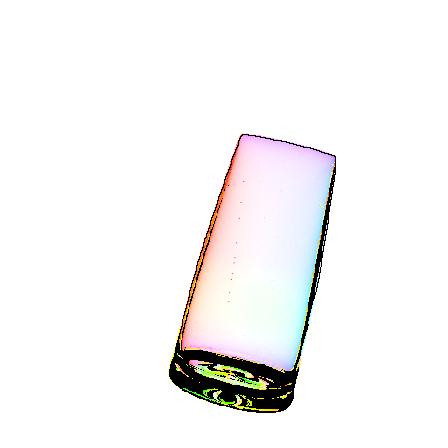}
    \includegraphics[width=0.095\textwidth]{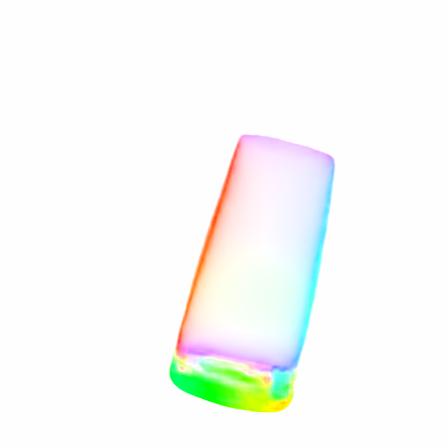}
    \includegraphics[width=0.095\textwidth]{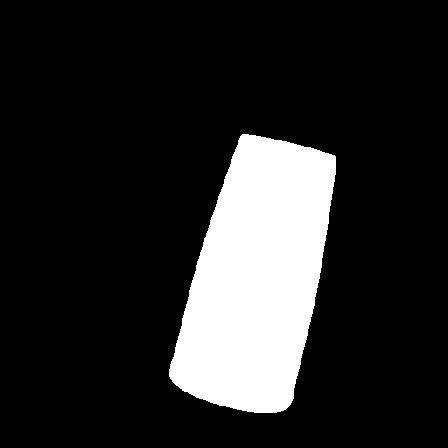}
    \includegraphics[width=0.095\textwidth]{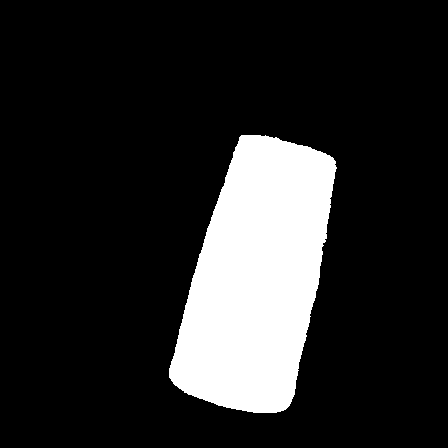}
    \includegraphics[width=0.095\textwidth]{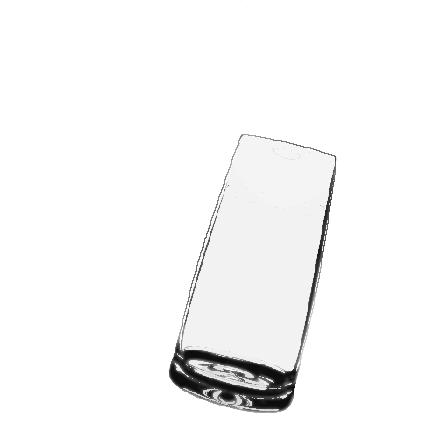}
    \includegraphics[width=0.095\textwidth]{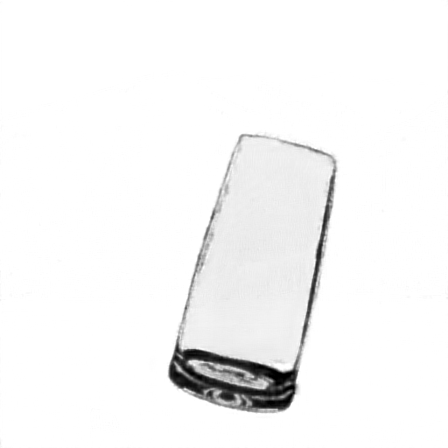}
    \\
    \vspace{-0.5em}\makebox[0.38\textwidth]{\scriptsize (a) Glass, I-MSE = $0.21 \times 10^{-2}$} 
    \makebox[0.19\textwidth]{\scriptsize F-EPE = 2.6 / 15.0} 
    \makebox[0.19\textwidth]{\scriptsize M-IoU = 0.99} 
    \makebox[0.19\textwidth]{\scriptsize A-MSE = $0.16 \times 10^{-2}$} 
    \\
    \includegraphics[width=0.095\textwidth]{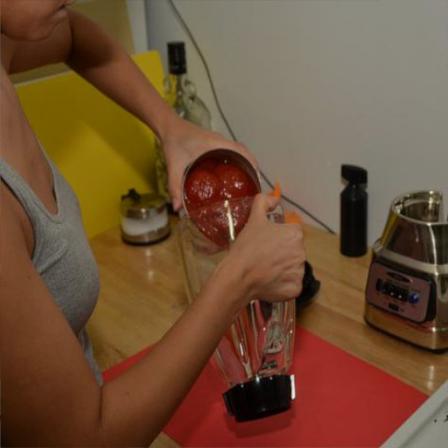}
    \includegraphics[width=0.095\textwidth]{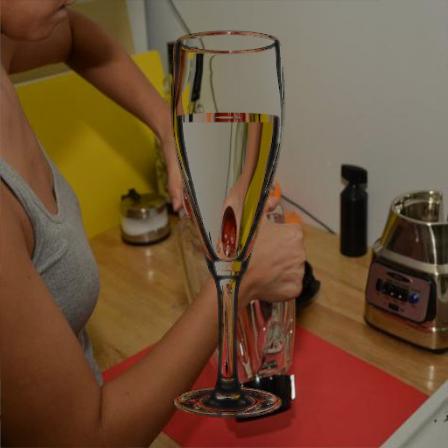}
    \includegraphics[width=0.095\textwidth]{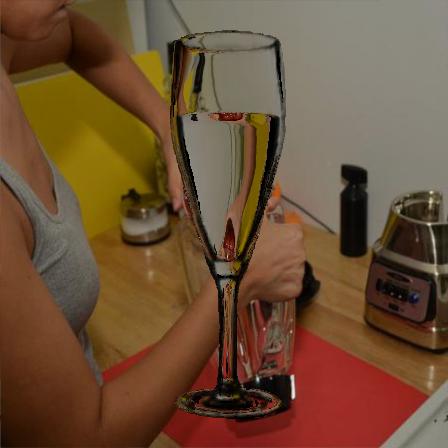}
    \includegraphics[width=0.095\textwidth]{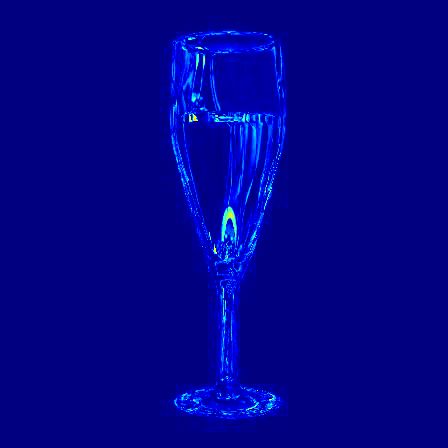}
    \includegraphics[width=0.095\textwidth]{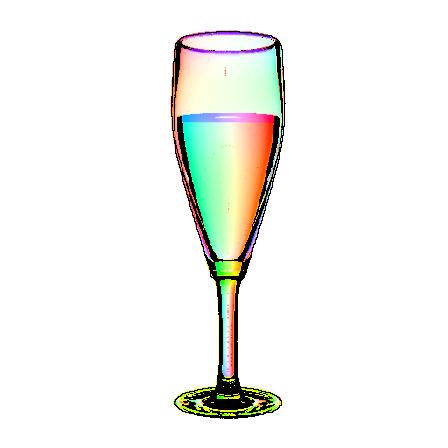}
    \includegraphics[width=0.095\textwidth]{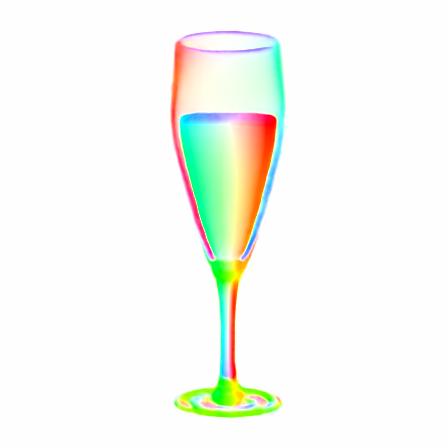}
    \includegraphics[width=0.095\textwidth]{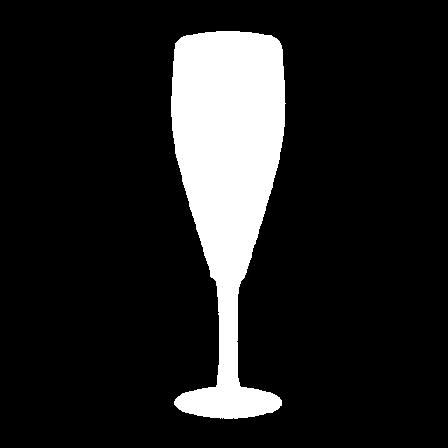}
    \includegraphics[width=0.095\textwidth]{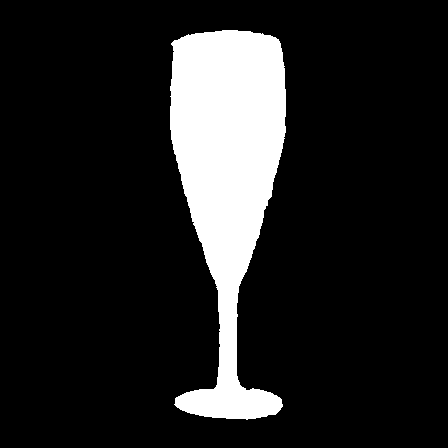}
    \includegraphics[width=0.095\textwidth]{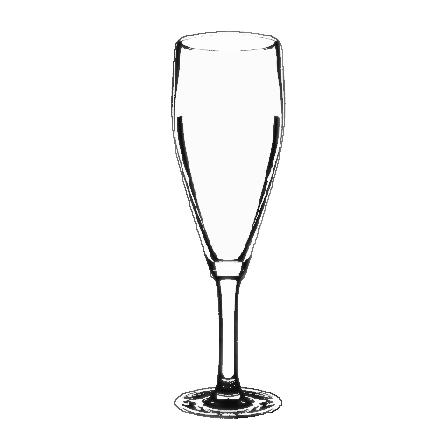}
    \includegraphics[width=0.095\textwidth]{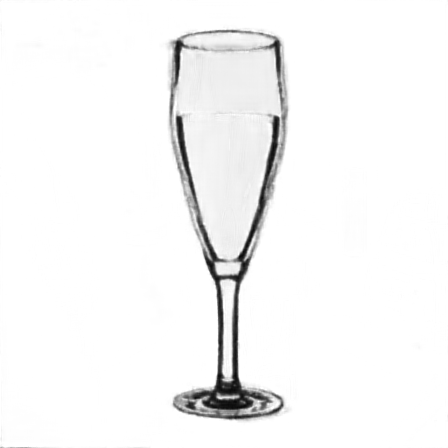}
    \\
    \vspace{-0.5em}\makebox[0.38\textwidth]{\scriptsize (b) Glass with Water, I-MSE = $0.15\times 10^{-2}$} 
    \makebox[0.19\textwidth]{\scriptsize F-EPE = 3.8 / 25.0} 
    \makebox[0.19\textwidth]{\scriptsize M-IoU = 0.97} 
    \makebox[0.19\textwidth]{\scriptsize A-MSE = 0.40 $\times 10^{-2}$} 
    \\
    \includegraphics[width=0.095\textwidth]{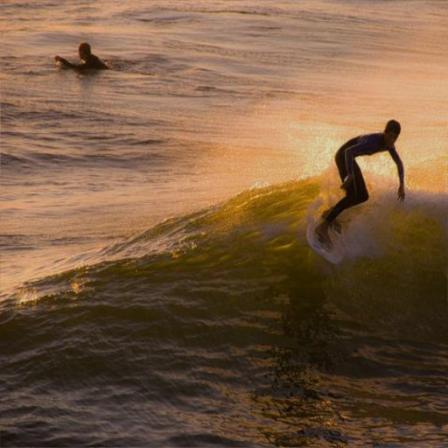}
    \includegraphics[width=0.095\textwidth]{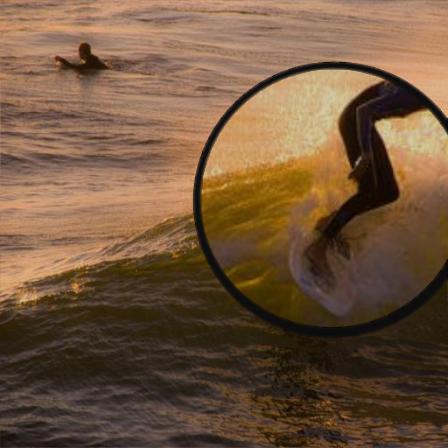}
    \includegraphics[width=0.095\textwidth]{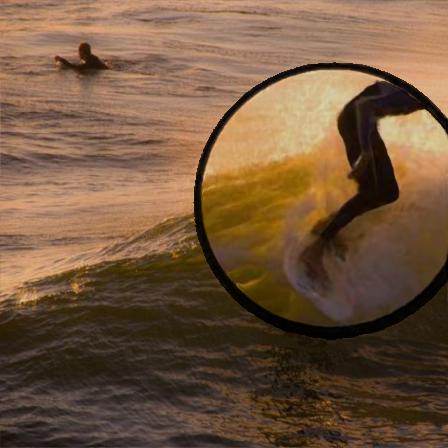}
   \includegraphics[width=0.095\textwidth]{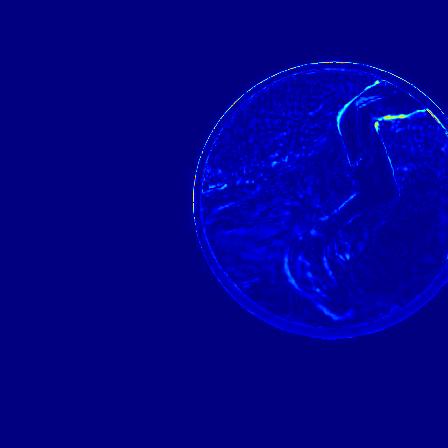}
    \includegraphics[width=0.095\textwidth]{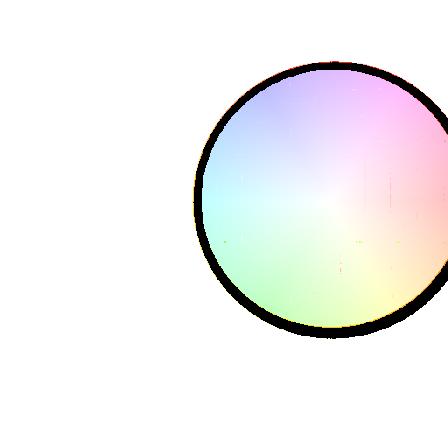}
    \includegraphics[width=0.095\textwidth]{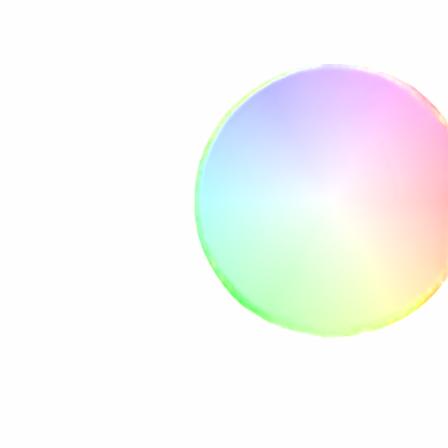}
    \includegraphics[width=0.095\textwidth]{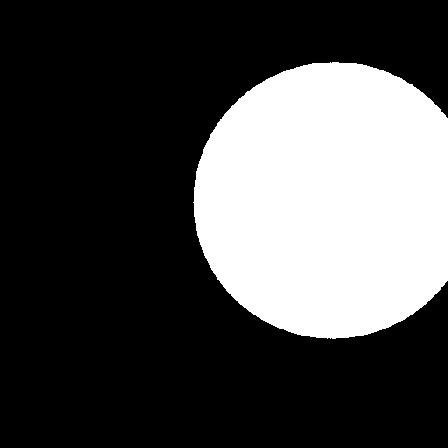}
    \includegraphics[width=0.095\textwidth]{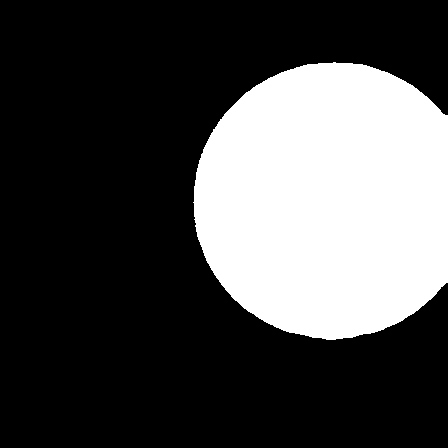}
    \includegraphics[width=0.095\textwidth]{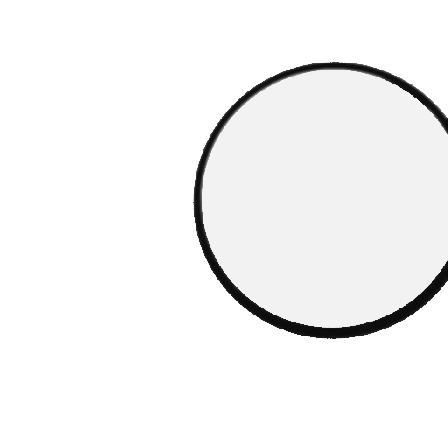}
    \includegraphics[width=0.095\textwidth]{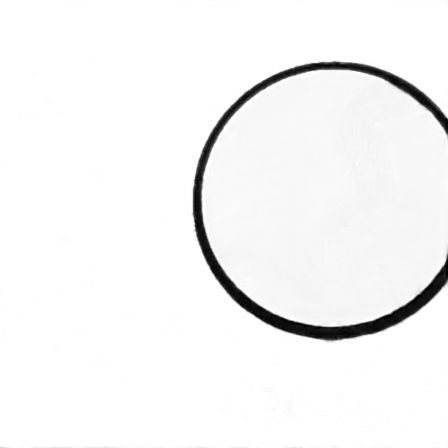}
    \\
    \vspace{-0.5em}\makebox[0.38\textwidth]{\scriptsize (c) Lens, I-MSE = $0.079\times 10^{-2}$}
    \makebox[0.19\textwidth]{\scriptsize F-EPE = 1.5 / 3.7} 
    \makebox[0.19\textwidth]{\scriptsize M-IoU = 1.00} 
    \makebox[0.19\textwidth]{\scriptsize A-MSE = $0.17\times 10^{-2}$} 
    \\
    \includegraphics[width=0.095\textwidth]{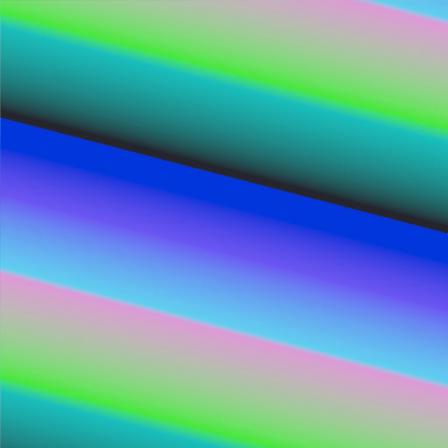}
    \includegraphics[width=0.095\textwidth]{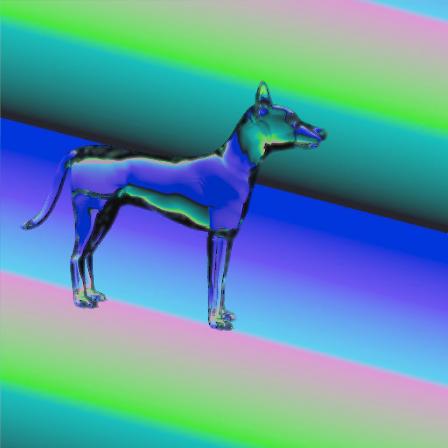}
    \includegraphics[width=0.095\textwidth]{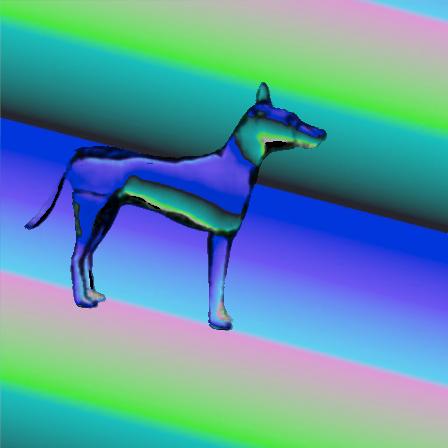}
    \includegraphics[width=0.095\textwidth]{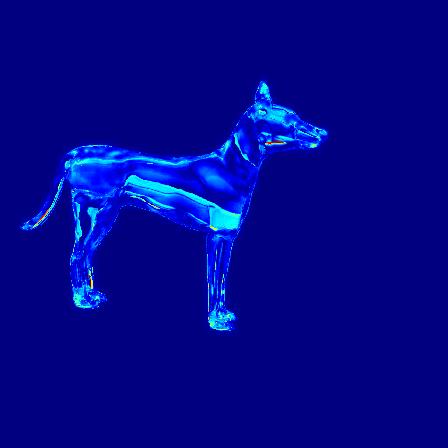}
    \includegraphics[width=0.095\textwidth]{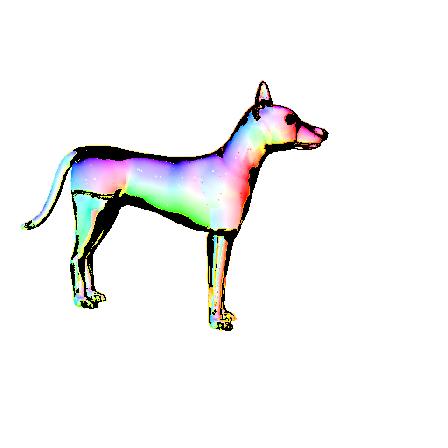}
    \includegraphics[width=0.095\textwidth]{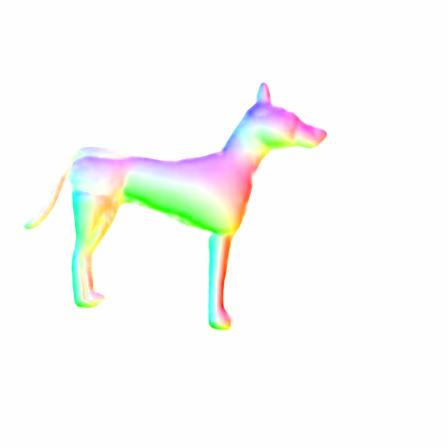}
    \includegraphics[width=0.095\textwidth]{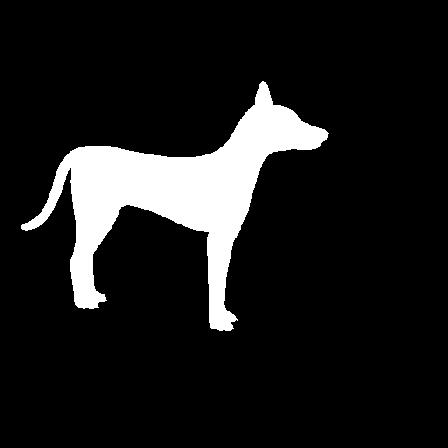}
    \includegraphics[width=0.095\textwidth]{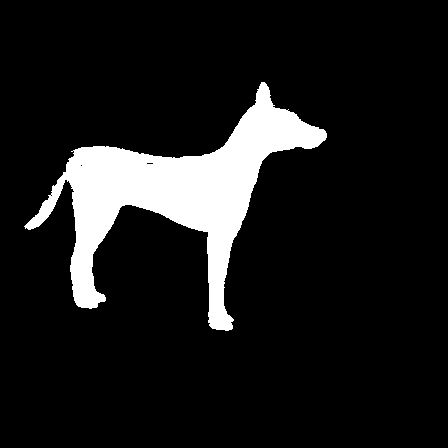}
    \includegraphics[width=0.095\textwidth]{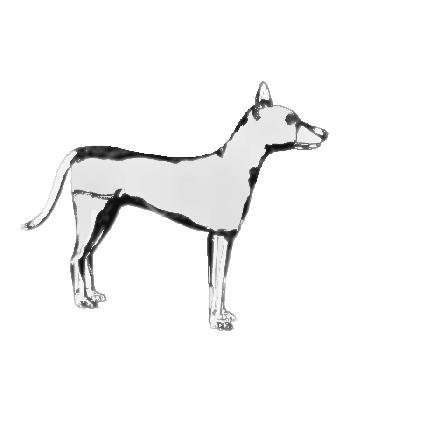}
    \includegraphics[width=0.095\textwidth]{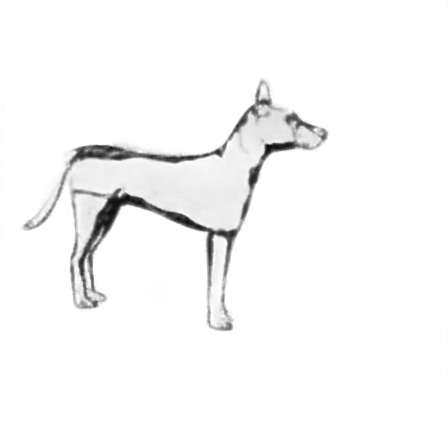}
    \\
    \vspace{-0.5em}\makebox[0.38\textwidth]{\scriptsize (d) Complex Dog, I-MSE = $0.28\times 10^{-2}$} 
    \makebox[0.19\textwidth]{\scriptsize F-EPE = 5.05 / 40.6} 
    \makebox[0.19\textwidth]{\scriptsize M-IoU = 0.96} 
    \makebox[0.19\textwidth]{\scriptsize A-MSE = $0.16\times 10^{-2}$} 
    \caption{Qualitative results on synthetic data. The first to the fourth columns show background, input image, reconstructed image, and reconstruction error map, respectively. Quantitative results are shown below each example. Dark region in GT flow indicates no valid flow. (Best viewed in PDF with zoom.)} \label{fig:qual_synth}
\end{figure*}

\subsection{Ablation Study for Network Structure}
\label{sub:Network Analysis}
We quantitatively analyzed different components of TOM-Net using synthetic dataset\footnote{Complex shape is excluded in experiments here to speed up training.}. In particular, we verified the effectiveness of \emph{image reconstruction loss} ($\mathcal{L}^{c}_{ir}$), \emph{refractive flow regression loss} ($\mathcal{L}^{c}_{fr}$), \emph{multi-scale loss}, \emph{cross-link}, and \emph{RefineNet}, where the first four components were evaluated by removing each of them from CoarseNet. Each variant was trained separately. RefineNet was evaluated by adding it to a trained CoarseNet and was trained while fixing the parameters of CoarseNet. Besides, we included a naive baseline, denoted as \emph{Background}, by considering a zero matte case (i.e., whole image as object mask, no attenuation, and no refractive flow) where the reconstructed image is the same as the background image. We evaluated end-point error (EPE) for refractive flow fields, intersection over union (IoU) for object masks, mean square error (MSE) for attenuation masks and image reconstruction results, respectively. The results are summarized in Tab. \ref{tab:self_compare}.  \emph{Background} was outperformed by all TOM-Net variants with a large margin for all the evaluation metrics, which clearly shows that TOM-Net can successfully learn the matte. Removing each component from CoarseNet, the overall performance decreased, although some metrics slightly increased due to learning trade-offs, demonstrating these four components are essential for TOM-Net.
By introducing RefineNet, the refractive flow performance was boosted, which verified the effectiveness of RefineNet (see Fig. \ref{fig:refine}). 

\subsection{Results on Synthetic Data}
\label{sub:Results on Synthetic data}
Quantitative results for synthetic validation dataset are presented in Tab. \ref{tab:quant_synth}. We compared TOM-Net against \emph{Background} and CoarseNet. Here, to accelerate training convergence, we first trained CoarseNet from scratch using our synthetic dataset excluding the complex shape subset. The trained CoarseNet was then fine-tuned using the entire training set including complex shapes, followed by training of RefineNet on the entire training set with random initialization. Similar to previous experiments, TOM-Net outperformed \emph{Background} with a large margin, and slightly outperformed CoarseNet in EPE and MSE, which implies more local details can be learned by RefinedNet. The errors of complex shape category are larger than that of others, because complex shapes contain more sharp regions that will induce more errors. Although TOM-Net is not expected to learn highly accurate refractive flow, the average EPE errors ($2.7/18.6$)\footnote{The first value is measured on the whole image and the second measured within the object region.} are very small compared with the dimensionality of the input image ($448\times 448$). In this sense, our predicted flow is capable of producing visually plausible refractive effect (see Fig. \ref{fig:qual_synth}). Although the background images and objects in the validation set never appear in the training set, TOM-Net can still predict robust matte. The pleasing results of the complex shapes also demonstrate that our model can generalize well from basic shapes to complex shapes.

\begin{table}
    \centering
    \caption{Quantitative results on real data.}
    \resizebox{0.48\textwidth}{!}{ 
    \huge
    \begin{tabular}{c|*{2}{c}|*{2}{c}|*{2}{c}|*{2}{c}|*{2}{c}}
        \toprule
        \multirow{2}{*}{} & \multicolumn{2}{c}{Glass} 
                          & \multicolumn{2}{c}{G \& W} 
                          & \multicolumn{2}{c}{Lens} 
                          & \multicolumn{2}{c}{Cplx} 
                          & \multicolumn{2}{c}{Avg} \\
        & PSNR & SSIM  & PSNR & SSIM & PSNR & SSIM & PSNR & SSIM & PSNR & SSIM \\
        \midrule
        Background    & 22.05 & 0.894 & 20.75 & 0.886 & 18.60 & 0.860 & 16.85 & 0.816 & 19.56 & 0.864 \\ 
        CoarseNet     & 25.09 & 0.921 & 23.53 & 0.911 & 21.13 & 0.895 & 17.89 & 0.835 & 21.91 & 0.891  \\ 
        TOM-Net        & 25.06 & 0.920 & 23.53 & 0.911 & 20.89 & 0.893 & 17.88 & 0.835 & 21.84 & 0.890 \\ 
        \bottomrule
    \end{tabular}
    }
    \label{tab:real_quant}
\end{table}

\begin{figure*} \centering
    \makebox[0.12\textwidth]{\footnotesize Background} 
    \makebox[0.12\textwidth]{\footnotesize Input} 
    \makebox[0.12\textwidth]{\footnotesize Rec. Image} 
    \makebox[0.12\textwidth]{\footnotesize Rec. Error} 
    \makebox[0.12\textwidth]{\footnotesize Ref. Flow} 
    \makebox[0.12\textwidth]{\footnotesize Obj. Mask} 
    \makebox[0.12\textwidth]{\footnotesize Att. Mask} 
    \makebox[0.12\textwidth]{\footnotesize Composite} 
    \\
    \raisebox{0.5\height}{\rotatebox{90}{\scriptsize (a) Glass}}
    \includegraphics[width=0.117\textwidth]{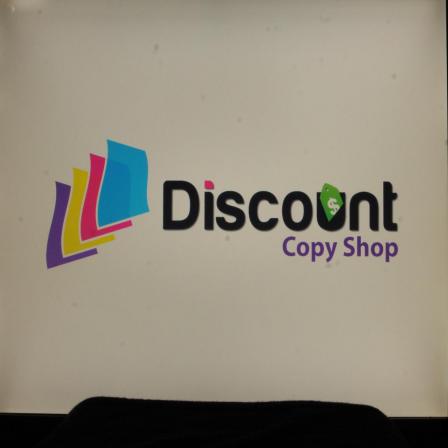}
    \includegraphics[width=0.117\textwidth]{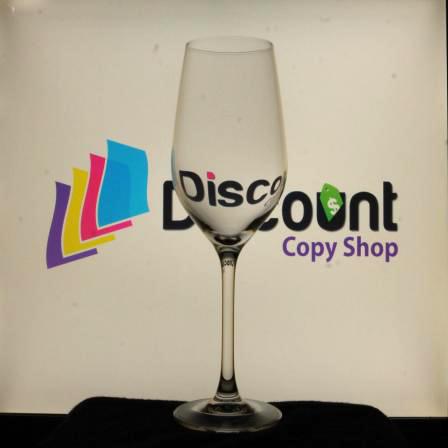}
    \includegraphics[width=0.117\textwidth]{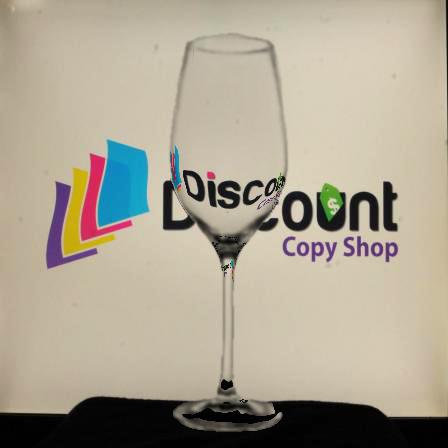}
    \includegraphics[width=0.117\textwidth]{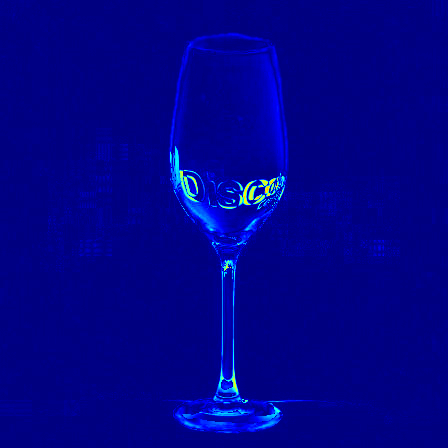}
    \raisebox{0.3\height}{\rotatebox{90}{\tiny P = 27.6, S = 0.95}}
    \includegraphics[width=0.117\textwidth]{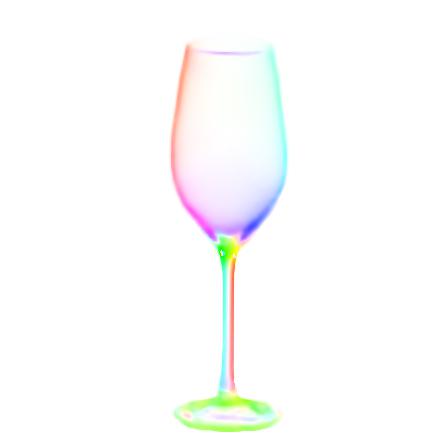}
    \includegraphics[width=0.117\textwidth]{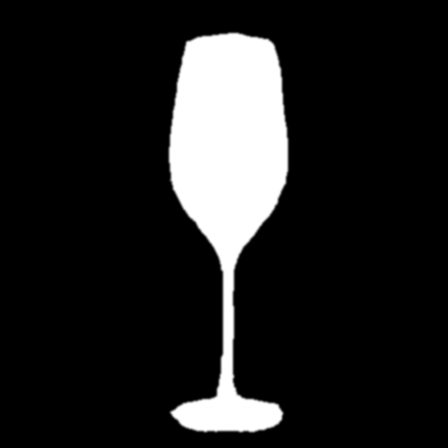}
    \includegraphics[width=0.117\textwidth]{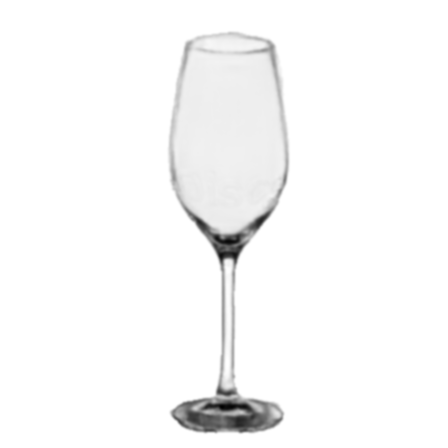}
    \includegraphics[width=0.117\textwidth]{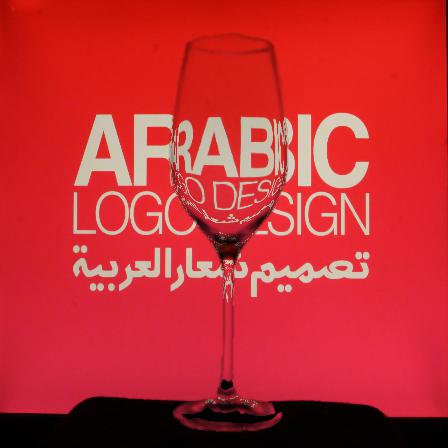}
    \\
    \raisebox{0.1\height}{\rotatebox{90}{\scriptsize (b) Glass \& Water}}
    \includegraphics[width=0.117\textwidth]{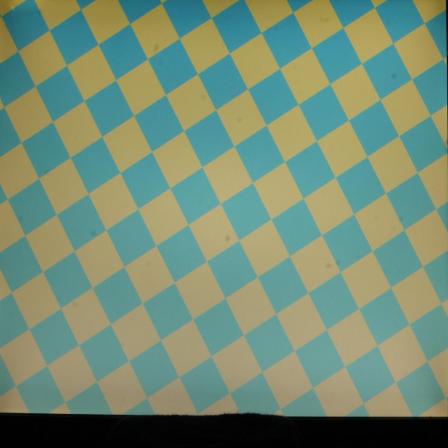}
    \includegraphics[width=0.117\textwidth]{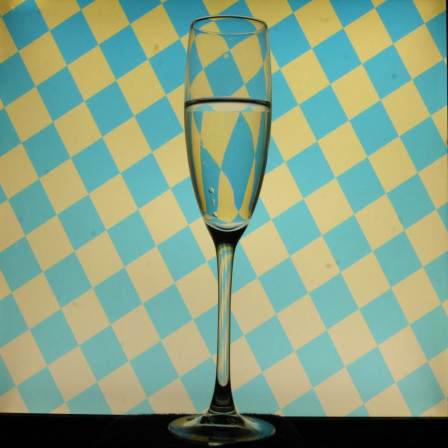}
    \includegraphics[width=0.117\textwidth]{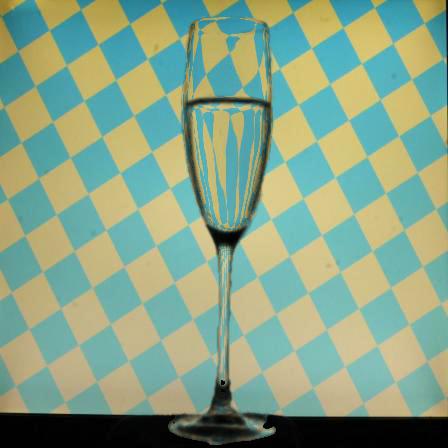}
    \includegraphics[width=0.117\textwidth]{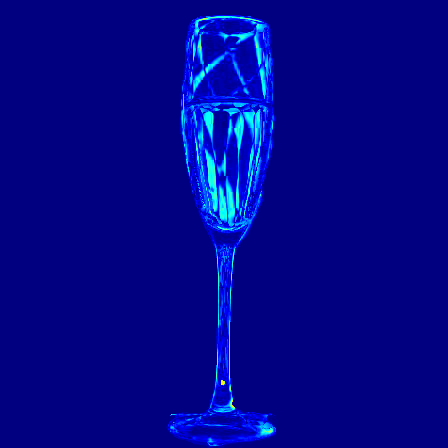}
    \raisebox{0.3\height}{\rotatebox{90}{\tiny P = 27.3, S = 0.95}}
    \includegraphics[width=0.117\textwidth]{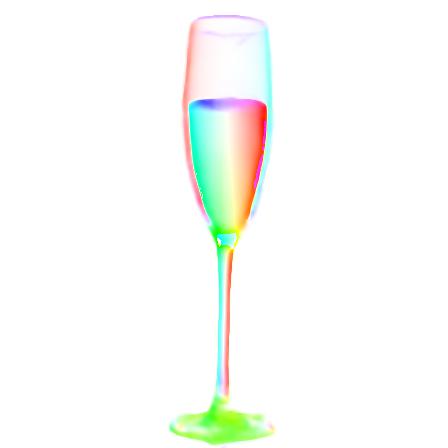}
    \includegraphics[width=0.117\textwidth]{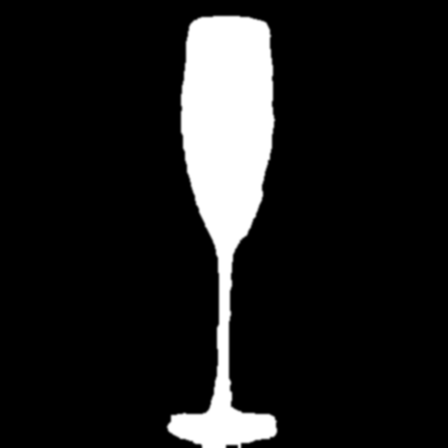}
    \includegraphics[width=0.117\textwidth]{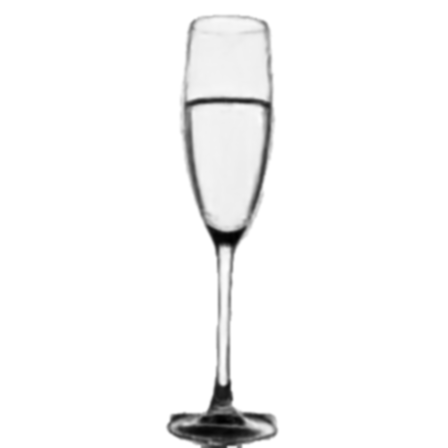}
    \includegraphics[width=0.117\textwidth]{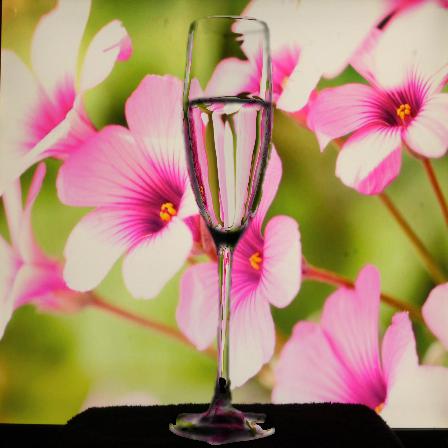}
    \\
    \raisebox{0.5\height}{\rotatebox{90}{\scriptsize (c) Lens}}
    \includegraphics[width=0.117\textwidth]{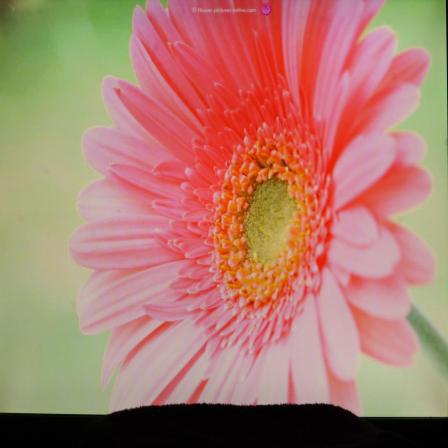}
    \includegraphics[width=0.117\textwidth]{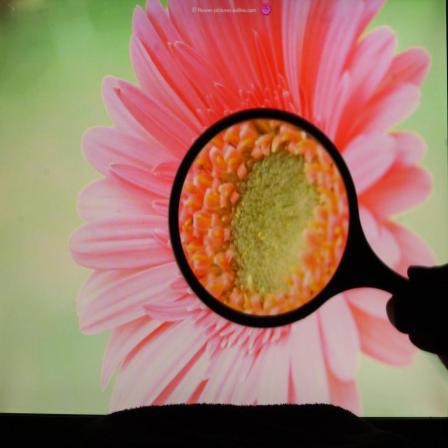}
    \includegraphics[width=0.117\textwidth]{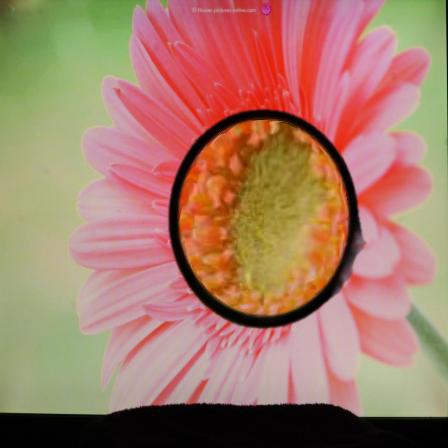}
    \includegraphics[width=0.117\textwidth]{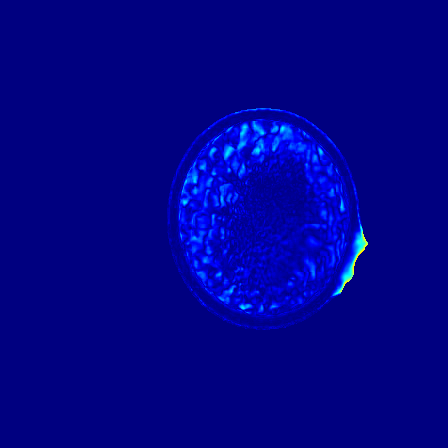}
    \raisebox{0.3\height}{\rotatebox{90}{\tiny P = 27.2, S = 0.91}}   
    \includegraphics[width=0.117\textwidth]{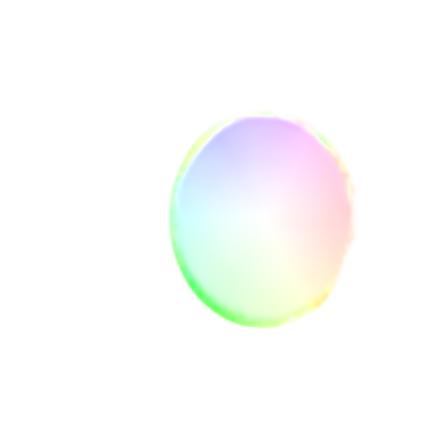}
    \includegraphics[width=0.117\textwidth]{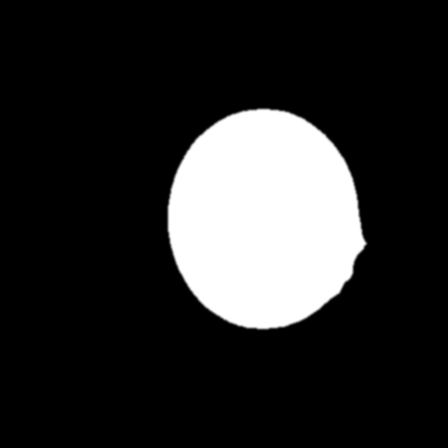}
    \includegraphics[width=0.117\textwidth]{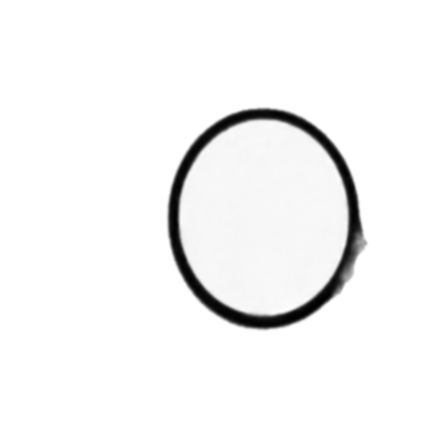}
    \includegraphics[width=0.117\textwidth]{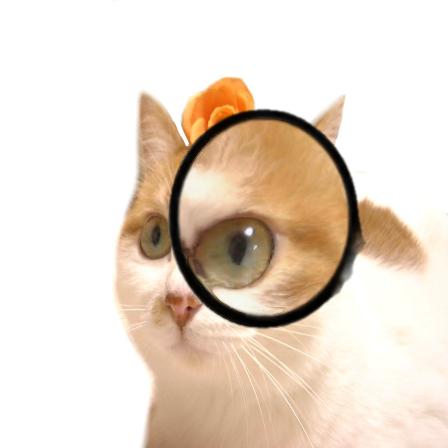}
    \\
    \raisebox{0.1\height}{\rotatebox{90}{\scriptsize (d) Multi-objects}}
    \includegraphics[width=0.117\textwidth]{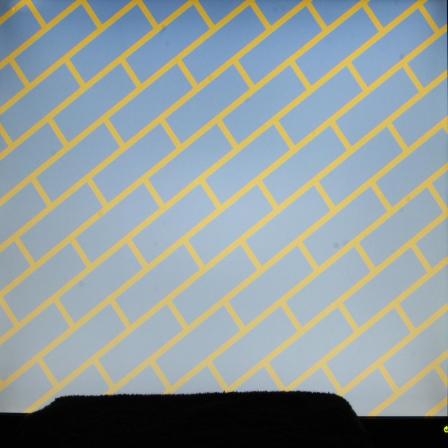}
    \includegraphics[width=0.117\textwidth]{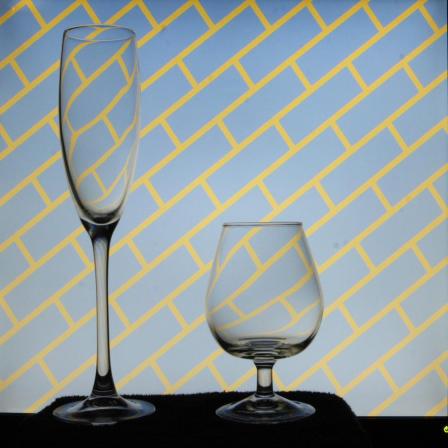}
    \includegraphics[width=0.117\textwidth]{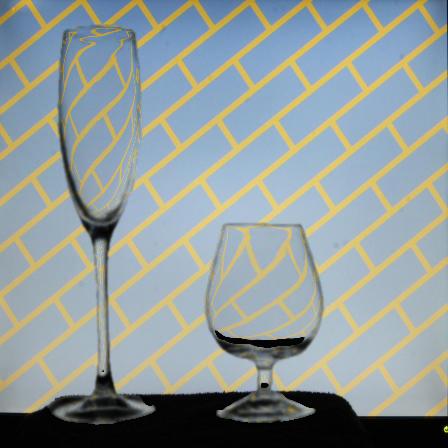}
    \includegraphics[width=0.117\textwidth]{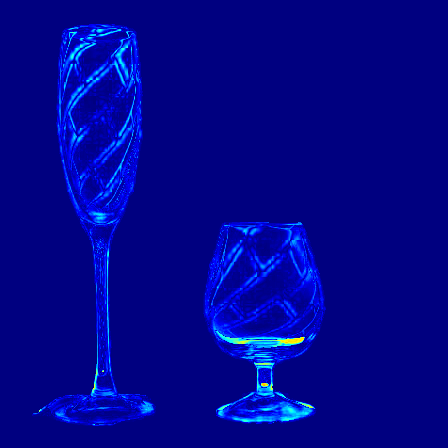}
    \raisebox{0.3\height}{\rotatebox{90}{\tiny P = 25.37, S = 0.93}}
    \includegraphics[width=0.117\textwidth]{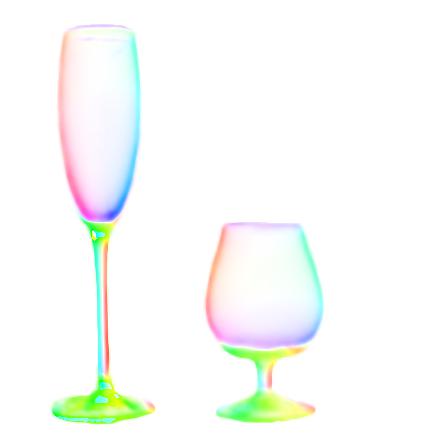}
    \includegraphics[width=0.117\textwidth]{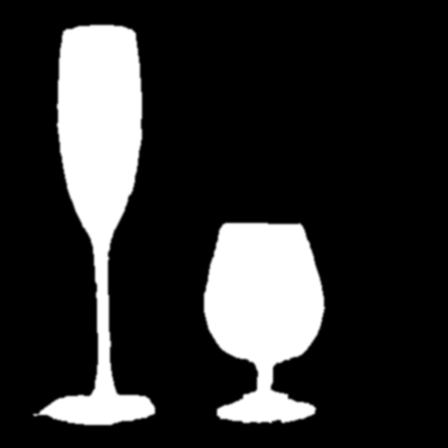}
    \includegraphics[width=0.117\textwidth]{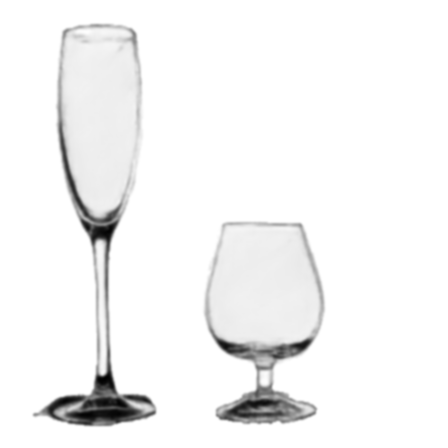}
    \includegraphics[width=0.117\textwidth]{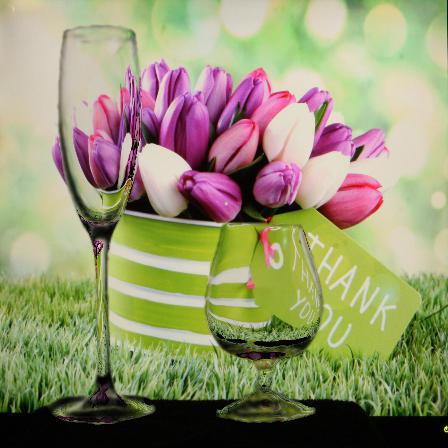}
    \\
    \raisebox{0.1\height}{\rotatebox{90}{\scriptsize (e) Complex Bull}}
    \includegraphics[width=0.117\textwidth]{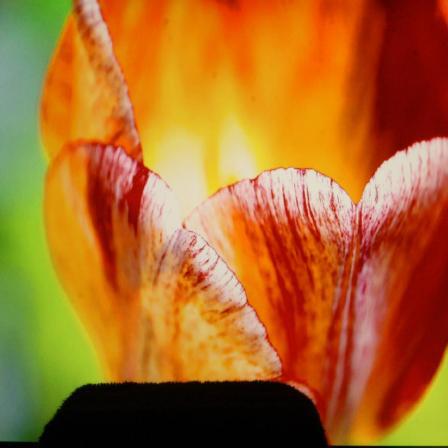}
    \includegraphics[width=0.117\textwidth]{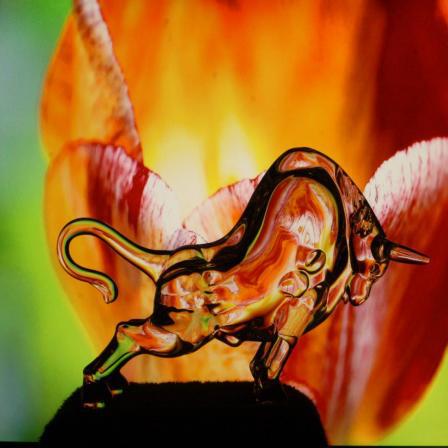}
    \includegraphics[width=0.117\textwidth]{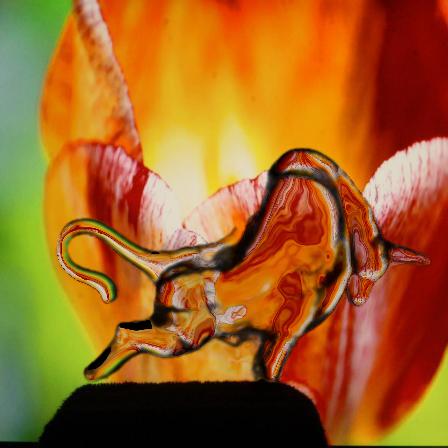}
    \includegraphics[width=0.117\textwidth]{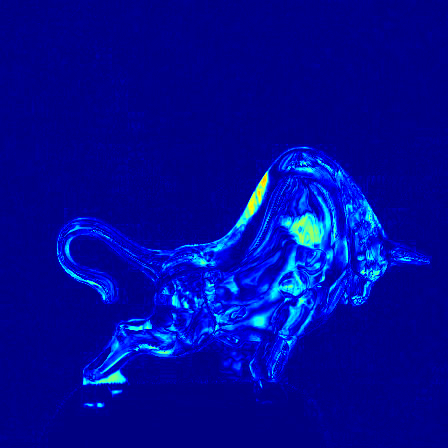}
    \raisebox{0.3\height}{\rotatebox{90}{\tiny P = 20.31, S = 0.84}}
    \includegraphics[width=0.117\textwidth]{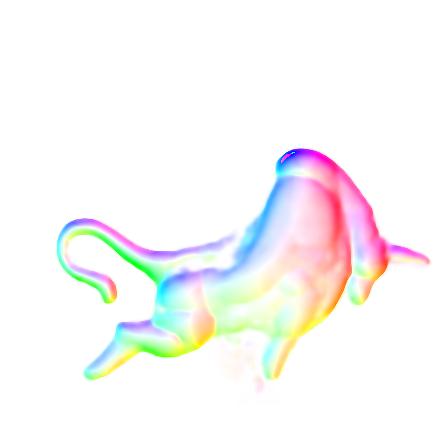}
    \includegraphics[width=0.117\textwidth]{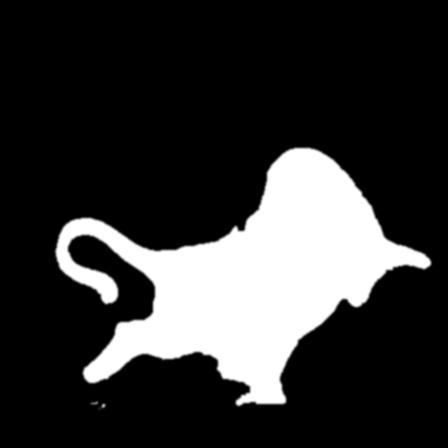}
    \includegraphics[width=0.117\textwidth]{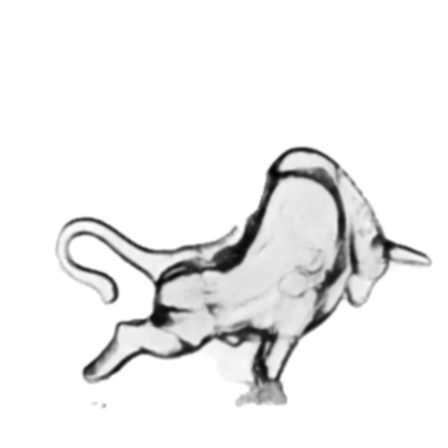}
    \includegraphics[width=0.117\textwidth]{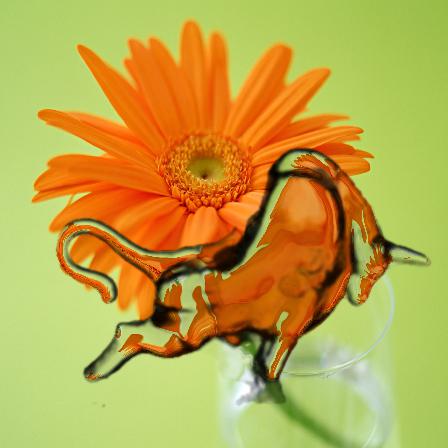}
    \\                                                                          
    \raisebox{0.\height}{\rotatebox{90}{\scriptsize (f) Complex Dragon}}
    \includegraphics[width=0.117\textwidth]{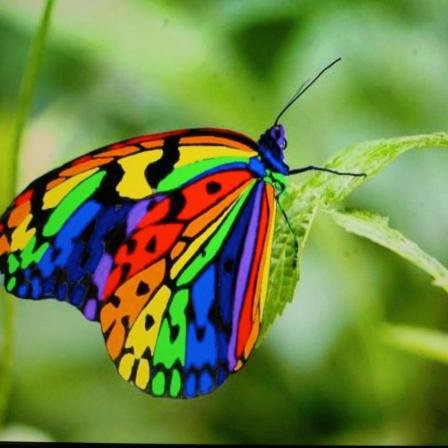}
    \includegraphics[width=0.117\textwidth]{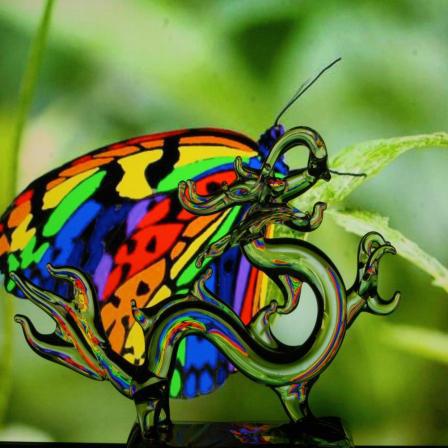}
    \includegraphics[width=0.117\textwidth]{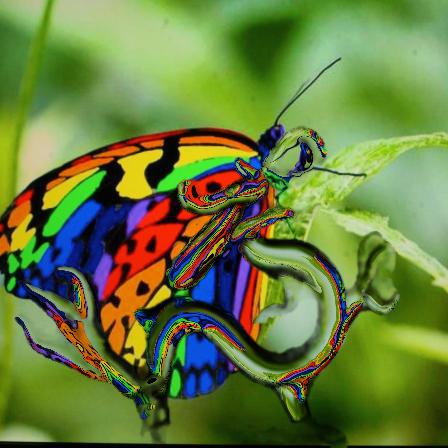}
    \includegraphics[width=0.117\textwidth]{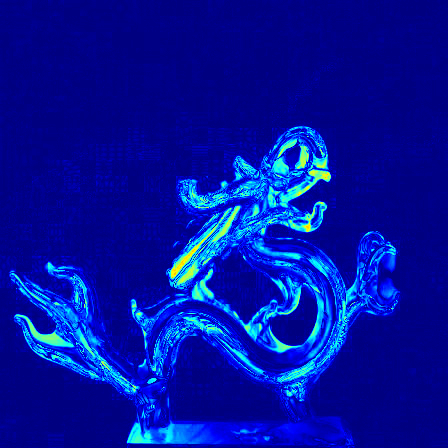}
    \raisebox{0.3\height}{\rotatebox{90}{\tiny P = 18.46, S = 0.80}}                   
    \includegraphics[width=0.117\textwidth]{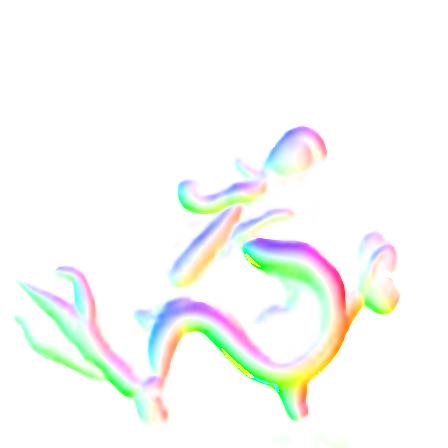}
    \includegraphics[width=0.117\textwidth]{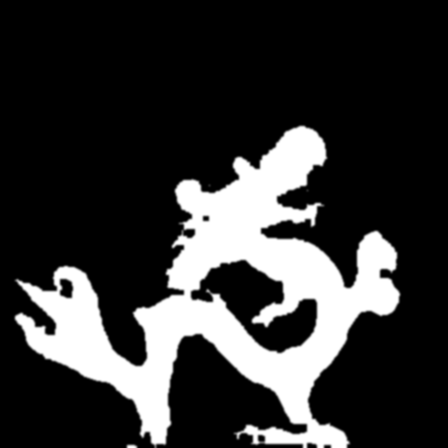}
    \includegraphics[width=0.117\textwidth]{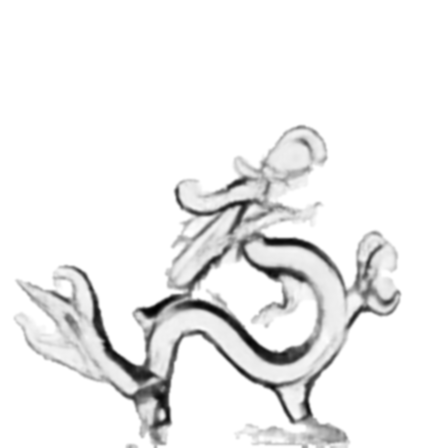}
    \includegraphics[width=0.117\textwidth]{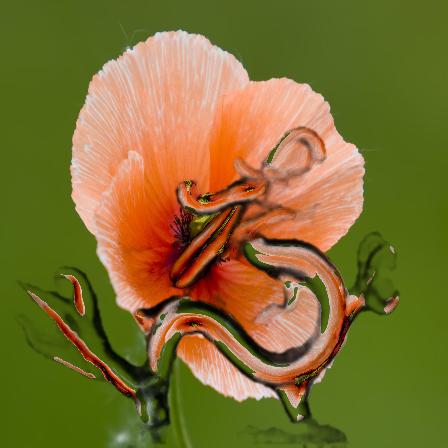}
    \\
    \caption{Qualitative results on real data. The PSNR and SSIM between input photographs and reconstructed images are shown right after the error maps. The last column shows the composites on novel backgrounds given the estimated matte. (Best viewed in PDF with zoom.)} \label{fig:real_qualitative}
\end{figure*}

\vspace{-0.2em} 
\subsection{Results on Real Data}
\label{sub:Results}
We evaluated TOM-Net on our collected real dataset, which consists of 876 images of real objects. Due to the absence of ground-truth matte, evaluation on the absolute error with ground truth is not possible. Instead, we evaluated PSNR and SSIM metrics \cite{wang2004image} between each pair of photograph and reconstructed image. The results are shown in Tab. \ref{tab:real_quant}. The average PSNR and SSIM are above $21.0$ and $0.89$. The values are a bit lower for complex shapes, due to the opaque base of complex objects as well as the sharp regions of the objects that might induce large errors. After training, TOM-Net generalized well to common real transparent objects (see Fig. \ref{fig:real_qualitative}). It is worth to note that during training, each sample contains only one object, while TOM-Net can predict reliable matte for images containing multiple objects, which indicates the transferability and robustness of TOM-Net.

\begin{table} \centering
    \caption{User study results.}
    \huge
    \resizebox{0.48\textwidth}{!}{
        \begin{tabular}{c|*{3}{c}|*{3}{c}|*{3}{c}|*{3}{c}|*{3}{c}}
        \toprule
        \multirow{2}{*}{} & \multicolumn{3}{c}{Glass} 
                               & \multicolumn{3}{c}{G \& W} 
                               & \multicolumn{3}{c}{Lens} 
                               & \multicolumn{3}{c}{Cplx}  
                               & \multicolumn{3}{c}{All}  \\
                               & P & C & N 
                               & P & C & N
                               & P & C & N 
                               & P & C & N 
                               & P & C & N \\
        \midrule
        Photographs        & 522 & 275 & 31 & 163 & 97 & 16 & 74 & 48 & 16 & 91 & 35 & 12 & 850 & 455 & 75 \\
        Composites    & 531 & 266 & 31 & 145 & 113 & 18 & 73 & 52 & 13 & 78 & 51 & 9 & 827 & 482 & 71 \\
        \bottomrule
    \end{tabular}
    }
    \label{tab:user_study}
\end{table}

\vspace{-1.3em} 
\paragraph{User Study}
\label{par:User Study}
A user study was carried out to validate the realism of TOM-Net composites. 69 subjects participated in our user study. At the beginning, we showed each participant photographs of the transparent objects that will be seen during the user study. The objects consisted of 3 different glasses, 1 glass with water, 1 lens, and 1 complex shape. 40 samples, including 20 photographs\footnote{glass $\times$12, glass \& water $\times$4, lens $\times$2, and complex shape $\times$2.} and the corresponding 20 TOM-Net composites, were then randomly presented to each subject. When showing each sample, we also showed the corresponding background image to the subject for reference. We provided 3 options for each sample: (P) {\em photograph}, (C) {\em composite}, (N) {\em not distinguishable}.
Tab. \ref{tab:user_study} shows the statistics of the user study. The 69 participants produced 1380 votes for the 20 real photographs, and 1380 votes for the 20 composites, respectively. The P:C:N ratios are $850:455:75$ and $827:482:71$ for photographs and composites respectively. The per-category ratio also follows a similar trend, indicating close chance of photographs and composites to be considered real, which further demonstrates TOM-Net can produce realistic matte. 

\section{Conclusion and Discussion}
\label{sec:Conclusion}
We have introduced a simple and efficient model for transparent object matting, and proposed a CNN architecture, called TOM-Net, that takes a single image as input and predicts environment matte as an object mask, an attenuation mask, and a refractive flow field in a fast feed-forward pass. Besides, we created a large-scale synthetic dataset and a real dataset as a benchmark for learning transparent object matting. Promising results have been achieved on both synthetic and real data, which clearly demonstrated the feasibility and effectiveness of the proposed approach. Since our model assumes objects to be colorless and specular transparent, TOM-Net cannot be applied to colored transparent objects, translucent objects and transparent objects with multiple mapping (i.e., refraction and reflection happen simultaneously at a surface point). We consider exploring better models and architectures to handle these scenarios as our future work.
\paragraph{Acknowledgments}
This project is supported by a grant from the Research Grant Council of the Hong Kong (SAR), China, under the project HKU 718113E. We gratefully acknowledge the support of NVIDIA Corporation with the donation of the Titan X Pascal GPU used for this research. We thank Yiming Qian for help with the synthetic data rendering.
{\small
\bibliographystyle{ieee}
\bibliography{gychen}
}
\clearpage
\appendix
\onecolumn
\section{Details of Network Structure}
\label{sec:Network Structure}
We have proposed a two-stage deep learning framework, called TOM-Net, for learning transparent object matting. 
The first stage is a multi-scale encoder-decoder network (i.e. CoarseNet) that takes a single image as input and predicts a triple consisting of an object mask, an attenuation mask and a refractive flow field. Although the estimated object mask in the first stage is robust, the attenuation mask and refractive flow field lack local details.
The second stage is a residual network (i.e. RefineNet) that refines the coarse matte to achieve a sharper attenuation mask and a more detailed refractive flow field. 

\subsection{CoarseNet}
\label{sub:Encoder-Decoder Net}
\begin{table}[h]
    \centering
    \definecolor{Gray}{gray}{0.9}
	\resizebox{0.90\textwidth}{!}{
    \begin{tabular}{cc}
        \begin{tabular}[t]{|*{6}{l|}}
        \hline
        \multicolumn{6}{|c|}{\textbf{Encoder}} \\
        \hline
        \textbf{layer} & \textbf{k} & \textbf{s} & \textbf{chns} & \textbf{d-f} & \textbf{input} \\
        \hline 
        conv1   & 3     & 1     & 3/16    & 1    & Image \\
        conv1b  & 3     & 1     & 16/16   & 1    & conv1 \\ 
        conv2   & 3     & 2     & 16/16   & 2    & conv1b \\
        conv2b  & 3     & 1     & 16/16   & 2    & conv2 \\ 
        conv3   & 3     & 2     & 16/32   & 4    & conv2b \\
        conv3b  & 3     & 1     & 32/32   & 4    & conv3 \\ 
        conv4   & 3     & 2     & 32/64   & 8    & conv3b \\
        conv4b  & 3     & 1     & 64/64   & 8    & conv4 \\ 
        conv5   & 3     & 2     & 64/128  & 16   & conv4b \\
        conv5b  & 3     & 1     & 128/128 & 16   & conv5 \\ 
        conv6   & 3     & 2     & 128/256 & 32   & conv5b \\
        conv6b  & 3     & 1     & 256/256 & 32   & conv6 \\ 
        conv7   & 3     & 2     & 256/256 & 64   & conv6b \\
        conv7b  & 3     & 1     & 256/256 & 64   & conv7 \\
        \hline
        \end{tabular}
        \begin{tabular}[t]{|*{6}{l|}}
            \hline
            \multicolumn{6}{|c|}{\textbf{Decoder}} \\
            \hline
            \textbf{layer} & \textbf{k} & \textbf{s} & \textbf{chns} & \textbf{d-f} & \textbf{input} \\
            \hline 
            conv\_up7\_m & 3     & 1     & 256/256   & 32  & conv7b \\
            conv\_up7\_a & 3     & 1     & 256/256   & 32  & conv7b \\
            conv\_up7\_f & 3     & 1     & 256/256   & 32  & conv7b \\
            \hline 
            \multicolumn{6}{|c|}{conv\_up7=conv\_up7\_m+conv\_up7\_a+conv\_up7\_f} \\
            \hline
            conv\_up6\_m & 3     & 1     & 256/128   & 16  & conv\_up7+conv6b\\
            conv\_up6\_a & 3     & 1     & 256/128   & 16  & conv\_up7+conv6b\\
            conv\_up6\_f & 3     & 1     & 256/128   & 16  & conv\_up7+conv6b\\
            \hline 
            \multicolumn{6}{|c|}{conv\_up6=conv\_up6\_m+conv\_up6\_a+conv\_up6\_f} \\
            \hline
            conv\_up5\_m & 3     & 1     & 128/64    & 8   & conv\_up6+conv5b \\
            conv\_up5\_a & 3     & 1     & 128/64    & 8   & conv\_up6+conv5b \\
            conv\_up5\_f & 3     & 1     & 128/64    & 8   & conv\_up6+conv5b \\
            \hline
            \multicolumn{6}{|c|}{conv\_up5=conv\_up5\_m+conv\_up5\_a+conv\_up5\_f} \\
            \hline
            m\_4         & 3     & 1     & 128/2     & 8   & conv\_up5+conv4b \\
            a\_4         & 3     & 1     & 128/1     & 8   & conv\_up5+conv4b \\
            f\_4         & 3     & 1     & 128/2     & 8   & conv\_up5+conv4b \\
            conv\_up4\_m & 3     & 1     & 128/32    & 4   & conv\_up5+conv4b \\
            conv\_up4\_a & 3     & 1     & 128/32    & 4   & conv\_up5+conv4b \\
            conv\_up4\_f & 3     & 1     & 128/32    & 4   & conv\_up5+conv4b \\
            \hline
            \multicolumn{6}{|c|}{conv\_up4=conv\_up4\_m+conv\_up4\_a+conv\_up4\_f} \\
            \hline
            m\_3         & 3     & 1     & 69/2      & 4   & conv\_up4+conv3b+(m\_4$^{\times 2}$+a\_4$^{\times 2}$+a\_4$^{\times 2}$) \\
            a\_3         & 3     & 1     & 69/1      & 4   & conv\_up4+conv3b+(m\_4$^{\times 2}$+a\_4$^{\times 2}$+a\_4$^{\times 2}$) \\
            f\_3         & 3     & 1     & 69/2      & 4   & conv\_up4+conv3b+(m\_4$^{\times 2}$+a\_4$^{\times 2}$+a\_4$^{\times 2}$) \\
            conv\_up3\_m & 3     & 1     & 69/16     & 2   & conv\_up4+conv3b+(m\_4$^{\times 2}$+a\_4$^{\times 2}$+a\_4$^{\times 2}$) \\
            conv\_up3\_a & 3     & 1     & 69/16     & 2   & conv\_up4+conv3b+(m\_4$^{\times 2}$+a\_4$^{\times 2}$+a\_4$^{\times 2}$) \\
            conv\_up3\_f & 3     & 1     & 69/16     & 2   & conv\_up4+conv3b+(m\_4$^{\times 2}$+a\_4$^{\times 2}$+a\_4$^{\times 2}$) \\
            \hline
            \multicolumn{6}{|c|}{conv\_up3=conv\_up3\_m+conv\_up3\_a+conv\_up3\_f} \\
            \hline
            m\_2         & 3     & 1     & 37/2      & 2   & conv\_up3+conv2b+(m\_3$^{\times 2}$+a\_3$^{\times 2}$+a\_3$^{\times 2}$) \\
            a\_2         & 3     & 1     & 37/1      & 2   & conv\_up3+conv2b+(m\_3$^{\times 2}$+a\_3$^{\times 2}$+a\_3$^{\times 2}$) \\
            f\_2         & 3     & 1     & 37/2      & 2   & conv\_up3+conv2b+(m\_3$^{\times 2}$+a\_3$^{\times 2}$+a\_3$^{\times 2}$) \\
            conv\_up2\_m & 3     & 1     & 37/16     & 1   & conv\_up3+conv2b+(m\_3$^{\times 2}$+a\_3$^{\times 2}$+a\_3$^{\times 2}$) \\
            conv\_up2\_a & 3     & 1     & 37/16     & 1   & conv\_up3+conv2b+(m\_3$^{\times 2}$+a\_3$^{\times 2}$+a\_3$^{\times 2}$) \\
            conv\_up2\_f & 3     & 1     & 37/16     & 1   & conv\_up3+conv2b+(m\_3$^{\times 2}$+a\_3$^{\times 2}$+a\_3$^{\times 2}$) \\
            \hline
            \multicolumn{6}{|c|}{conv\_up2=conv\_up2\_m+conv\_up2\_a+conv\_up2\_f} \\
            \hline
            m\_1         & 3     & 1     & 37/2      & 1   & conv\_up2+conv1b+(m\_2$^{\times 2}$+a\_2$^{\times 2}$+a\_2$^{\times 2}$) \\
            a\_1         & 3     & 1     & 37/1      & 1   & conv\_up2+conv1b+(m\_2$^{\times 2}$+a\_2$^{\times 2}$+a\_2$^{\times 2}$) \\
            f\_1         & 3     & 1     & 37/2      & 1   & conv\_up2+conv1b+(m\_2$^{\times 2}$+a\_2$^{\times 2}$+a\_2$^{\times 2}$) \\
            \hline                                                                         
        \end{tabular}
    \end{tabular}
    }

    \caption{Network architecture of CoarseNet. \textbf{k} is the kernel size, \textbf{s} the stride, \textbf{chns} the number of input and output channels for each convolutional layer, \textbf{d-f} the down-sampling factor of the output for each layer relative to the input image, and \textbf{input} the input of each layer. 
    ``+" is a concatenation and ``$^{\times2}$" is a 2$\times$ nearest up-sampling operation. All convolutional layers are followed by BatchNorm and ReLU layers, except for the output layers. 
    For decoder, conv\_up$n$\_m is a convolutional layer in object mask branch followed by Batchnorm, ReLU and nearest up-sampling layers. Similarly, conv\_up$n$\_a and conv\_up$n$\_f are the layers in attenuation and refractive flow branch, respectively. m\_$n$, a\_$n$, f\_$n$ are the output of the CoarseNet, corresponding to the softmax normalized object mask probability, attenuation mask and refractive flow field of scale $n$.}
    \label{tab:CoarseNet}
\end{table}

\clearpage 
\subsection{RefineNet}
\label{sub:Encoder-Decoder Net}
RefineNet is a residual network. For memory and training efficiency, we first use three 2$\times$ down-sampling convolutional layers to reduce the spatial size of the input, followed by five residual blocks containing ten convolutional layers. To make the output of the same spatial size with the input, two up-sampling branches, each with three 2$\times$ de-convolutional layers, are used to regress a sharper attenuation mask and a more detailed refractive flow field. Tab. \ref{tab:refineNet} shows the details of the RefineNet.
\begin{table}[h]
    \centering
    \definecolor{Gray}{gray}{0.9}
	\resizebox{0.4\textwidth}{!}{
        \begin{tabular}[t]{|*{6}{l|}}
        \hline
        \multicolumn{6}{|c|}{\textbf{RefineNet}} \\
        \hline
        \textbf{layer} & \textbf{k} & \textbf{s} & \textbf{chns} & \textbf{d-f} & \textbf{input} \\
        \hline 
        conv1   & 9     & 1    & 8/64   & 1    & Image+m\_1+a\_1+f\_1\\
        conv2   & 4     & 2    & 64/64  & 2    & conv1 \\
        conv3   & 4     & 2    & 64/64  & 4    & conv2 \\
        conv4   & 4     & 2    & 64/64  & 8    & conv3 \\
        \hline 
        ResBlock1 & 3     & 1  & 64/64  & 8    & conv4 \\
        ResBlock2 & 3     & 1  & 64/64  & 8    & ResBlock1 \\
        ResBlock3 & 3     & 1  & 64/64  & 8    & ResBlock2 \\
        ResBlock4 & 3     & 1  & 64/64  & 8    & ResBlock3 \\
        ResBlock5 & 3     & 1  & 64/64  & 8    & ResBlock4 \\
        \hline 
        deconv1\_a  & 4   & 2    & 64/64 & 4   & ResBlock5 \\
        deconv2\_a  & 4   & 2    & 64/64 & 2   & deconv1\_a \\
        deconv3\_a  & 4   & 2    & 64/64 & 1   & deconv2\_a \\
        \rowcolor{gray!25}
        a\_refined  & 3   & 1    & 65/1 & 1   & deconv3\_a+a\_1\\
        \hline
        deconv1\_f  & 4   & 2    & 64/64 & 4   & ResBlock5 \\
        deconv2\_f  & 4   & 2    & 64/64 & 2   & deconv1\_f \\
        deconv3\_f  & 4   & 2    & 64/64 & 1   & deconv2\_f \\
        \rowcolor{gray!25}
        f\_refined        & 3   & 1    & 66/2 & 1   & deconv3\_f+f\_1\\
        \hline
        \end{tabular}
    }

    \caption{Network architecture of RefineNet. Input for conv1 layer is the concatenation of softmax normalized object mask probability (m\_1), attenuation (a\_1) and flow field (f\_1) from the output of CoarseNet. The final output of RefineNet are the refined attenuation (a\_refined) and the refined flow field (f\_refined). All convolutional layers are followed by BatchNorm and ReLU, again except for the output layers.}
    \label{tab:refineNet}
\end{table}

\end{document}